\documentclass{article}

\usepackage[preprint]{neurips_2025}

\usepackage[utf8]{inputenc} 
\usepackage[T1]{fontenc}    
\usepackage{hyperref}       
\usepackage{url}            
\usepackage{booktabs}       
\usepackage{amsfonts}       
\usepackage{nicefrac}       
\usepackage{microtype}      
\usepackage{xcolor}         

\usepackage{amsmath}
\usepackage{amssymb}
\usepackage{mathtools}
\usepackage{amsthm}
\usepackage[capitalize,noabbrev]{cleveref}

\theoremstyle{plain}

\theoremstyle{definition}

\theoremstyle{remark}

\newcommand{\defeq}{\mathrel{\mathop:}=}

\usepackage{dsfont}
\usepackage{comment}

\DeclareMathOperator*{\argmin}{arg\,min}

\usepackage{algorithm}
\usepackage{algorithmic}

\usepackage{tikz}

\title{Collision-based Dynamics for \\Multi-Marginal Optimal Transport}

\author{%
  Mohsen~Sadr\thanks{Correspondence to: \texttt{mohsen.sadr@icloud.com}} \\
Department of Mechanical Engineering, MIT, Cambridge,  USA\\
 Paul Scherrer Institute,  Villigen, Switzerland \\
  \texttt{mohsen.sadr@icloud.com} \\
  \And
  Hossein Gorji\\
  Laboratory for Computational Engineering, Empa, Dübendorf, Switzerland\\
  \texttt{mohammadhossein.gorji@empa.ch}
}

\begin{document}

\maketitle

\begin{abstract}
Inspired by the Boltzmann kinetics, we propose a collision-based dynamics with a Monte Carlo solution algorithm that approximates the solution of the multi-marginal optimal transport problem via randomized pairwise swapping of sample indices. The computational complexity and memory usage of the proposed method scale linearly with the number of samples, making it highly attractive for high-dimensional settings. In several examples, we demonstrate the efficiency of the proposed method compared to the state-of-the-art methods.
\end{abstract}

\section{Introduction}

\noindent 
Since its introduction by Gaspard Monge \cite{monge1781memoire} and seminal contributions by 
\cite{kantorovich1942translocation}, the Optimal Transport (OT) has evolved into a rich mathematical framework with fruitful theoretical properties. At its core, it gives a geometrically intuitive basis to compare and interpolate probability distributions, leading to wide-range of applications across many fields.
This includes interpolation between images \cite{ferradans2014regularized},  clustering dataset \cite{del2019robust}, surrogate models \cite{jacot2024empowering}, calibration of stochastic processes \cite{mohajerin2018data}, trajectory inference \cite{huguet2022manifold}, and finding the N-body particle distribution function in density functional theory \cite{cotar2013density} among other \cite{peyre2019computational}. However, the computational complexity associated with the underlying optimization problem limits the use of OT in large datasets. This is due to the fact that the OT problem is inherently linear programming over an infinite-dimensional space, resulting in computationally intensive optimization. The problem can even become intractable if multi-marginals are considered. Though non-exclusively, we can categorize the main computational algorithms for numerical solution to the OT problem as the following.
\\ \ \\
\noindent \textit{Linear programming.} This is the direct approach in solving the OT problem,  also known as Earth Mover's Distance in the literature \cite{pele2009fast}. Linear programming has been applied mainly to the two-marginal OT problem, where the computational complexity becomes $\mathcal{O}(N_p^3 \log(N_p))$ for $N_p$  number of samples per marginal. Fast EMD algorithms use the network simplex with empirical computational complexity of $\mathcal{O}(N_p^2)$ \cite{bonneel2011displacement}.
\\ \ \\
\noindent \textit{Regularization via entropy.} By incorporating entropy in the cost functional of the two-marginal OT, one derives a relaxed version of the OT problem \cite{cuturi2013sinkhorn,genevay2016stochastic}. The resulting optimization problem is convex and can be solved efficiently using the so-called Sinkhorn method, with the computational complexity of $\mathcal{O}(N_p^2)$.
\\ \ \\
\noindent \textit{Dynamic processes.}  Optimal maps can be described by dynamic processes. For example, the fluid dynamic formulation, given by Benamou-Brenier dynamics, describes the OT problem in the form of the Hamilton-Jacobi dynamics \cite{benamou2000computational}. Another class of dynamic formulation is given by marginal preserving processes, where OT (and its entropy regularized version) is recovered at the stationary state \cite{conforti2023projected}. In particular, Orthogonal Coupling Dynamics \cite{sadr2024ot} introduces an efficient algorithm with the computational complexity of $\mathcal{O}(N_p \log(N_p))$.
\\ \ \\
\noindent \textit{Reduction to assignment problem.}
As shown by \cite{ruschendorf1990characterization}, OT in discrete setting is closely related to the assignment problem. A notably efficient approximate solution is introduced by Iterative Swapping Algorithm (ISA) \cite{puccetti2017algorithm,puccetti2020computation}, where a near-optimal permutation of discrete points is found via consecutive swaps of samples in each marginal. 
The ISA has the computational complexity of $\mathcal{O}(N_p^2)$, and is the closest method to what we develop here. 
\\ \ \\
\noindent Other notable approximate methods include graph-based models \cite{haasler2021multi},  moment-based methods \cite{mula2024moment,sadr2024wasserstein}, simulated annealing \cite{ye2017simulated}, and sliced-Wasserstein \cite{bonneel2015sliced,huang2021riemannian}.
\\ \ \\
\noindent \textbf{Contributions.} 
We focus on the multi-marginal OT problem in the discrete setting, where only samples of the marginals are provided. Unlike the optimal assignment approach, which looks for the optimal permutation of samples, we iteratively improve the permutation of samples by pair-wise swapping. Instead of checking all possible swaps, which is pursued in ISA, we devise a random algorithm inspired by Boltzmann kinetics, where binary collisions are performed by randomly selecting collision pairs. We formalize both ISA and our collision-based algorithm, and motivate their consistency.    
\\ \ \\
We show that in the case of the OT problem with $L^p$-transport cost, the complexity of checking the condition for a swap to be accepted/rejected is independent of number of samples. Furthermore, we show that the complexity of the collision-based method scales linearly with number of samples. We empirically investigate its convergence behaviour and observe exponential convergence to the stationary solution, regardless of number of samples/marginals/dimension. In several toy examples, we show the error and performance of the proposed method compared to Sinkhorn and EMD. Then, as a show case, we assess the flexibility of the method in finding the optimal map in a five-marginal problem, which allows us to learn a map between normal and other target densities. As an application in Machine Learning, we demonstrate the performance of collision-based method in finding the distribution of the Wasserstein distance in the Japanese female facial expression, butterfly, and CelebA datasets.
\\ \ \\
\noindent \textbf{Definitions, Notations, and Problem Setup.} Let $\mathcal{P}(\mathcal{X}_i)$ be the space of non-negative Borel measures over $\mathcal{X}_i\subset \mathbb{R}^n$, and
\begin{eqnarray}
\mathcal{P}_2(\mathcal{X}_i)\defeq \left\{\mu\in\mathcal{P}(\mathcal{X}_i)\bigg\vert \int_{\mathcal{X}_i}\lVert x \rVert_2^2 \mu(dx)<\infty\right \} 
\end{eqnarray} 
with $\lVert . \rVert_2$  the usual $L^2$-Euclidean norm.
Consider $K$ probability measures $\mu_{i}\in \mathcal{P}_2(\mathcal{X}_i)$ with $i\in\{1,...,K\}$, and vanishing on $(n-1)$-rectifiable sets \cite{gangbo1998optimal}. We are interested in the Multi-Marginal Optimal Transport problem (MMOT), which seeks the minimization
\begin{eqnarray}
\label{eq:mm-ot-gen}
\pi_{\textrm{opt}}\defeq \argmin_{\pi\in \Pi(\mu_1,...,\mu_K)} \int_{\mathcal{X}} c(x_1,...,x_K) \pi(dx) \ ,
\end{eqnarray}
where $\mathcal{X}$ is the product set $\mathcal{X}\defeq \mathcal{X}_1 \times ... \times \mathcal{X}_K$. The optimization is constrained on $\Pi$, which is the set of coupling measures 
\begin{flalign}
\Pi(\mu_1,...,\mu_K)\defeq \Big\{\pi\in \mathcal{P}_2(\mathcal{X})&\bigg\vert \textrm{proj}_i(\pi)=\mu_i\ \forall i\in\{1,...,K\}\Big\} 
\end{flalign}
\sloppy and $\textrm{proj}_i: \mathcal{X}\to\mathcal{X}_i$ is the canonical projection. In order for MMOT to have a solution, it is sufficient to assume that the cost $c:\mathcal{X}\to \mathbb{R}$ is lower-semicontinuous \cite{gangbo1998optimal}. In general there is no guarantee that the optimal transport plan, i.e. $\pi_{\textrm{opt}}$, is induced by an optimal map. The existence of the optimal map entails further constraints on the cost. 
One interesting setting, which has been analyzed thoroughly,
is when $K=2$ and $c(x_1,x_2)=\lVert x_1-x_2\rVert_2^2$. For this $L^2$-OT setting, the optimal map exists and is unique \cite{gangbo1996geometry,caffarelli2017allocation}. 
The generalization of $L^2$-OT to MMOT has been carried out by the seminal work of Gangbo and Swiech \cite{gangbo1998optimal}. Consider 
\begin{eqnarray}
c(x_1,...,x_K)=\sum_{i=1}^K\sum_{j=i+1}^K \frac{1}{2}\lVert x_i-x_j \rVert_2^2 \ ,
\end{eqnarray}
the optimal plan $\pi_{\textrm{opt}}$ then takes a deterministic form and is concentrated on optimal maps. In fact, the equivalent form of \eqref{eq:mm-ot-gen} is given by optimization over the maps $\{T_i\}\in\mathcal{T}_K$, where
\begin{flalign}
\mathcal{T}_K\defeq\Big\{T=\{T_i\}_{i=1,..,K}\ \vert& \ \  T_i:\mathcal{X}_1\to\mathcal{X}_i,\ T_i\#\mu_1=\mu_i, \ T_1=\textrm{id}\Big\}
\end{flalign}
such that 
\begin{eqnarray}
\inf \Big\{\int_{\mathcal{X}_1}\sum_{i=1}^K\sum_{j=i+1}^K & \frac{1}{2}\lVert T_i(x_1)-T_j(x_1)\rVert_2^2 \ \mu_1(dx_1)\ \big \vert \  \{T_i\}_{i=1,...,K}\in \mathcal{T}_K\Big\}
\end{eqnarray}
is attained \cite{nenna2016numerical,gangbo1998optimal}. Though our devised algorithm is not restricted to a specific choice of $c(.)$, in order to keep the study focused, we target the Gangbo-Swiech setting and refer to it as $L^2$-MMOT problem. It is clear that the setting reduces to $L^2$-OT for $K=2$. We refer the reader to \cite{pass2015multi} and references therein for detailed analysis of existence and uniqueness in MMOT problem. 
\\ \ \\
Instead of direct access to $\{\mu_i\}$, we consider the scenario where only $N_p$ independent samples of each marginal, i.e. $\hat{X}^{(i)}=\{\hat{X}_1^{(i)},...,\hat{X}_{N_p}^{(i)}\}\sim \mu_i$, is known. Given $\{\hat{X}^{(i)}\}$ for $i\in\{1,...,K\}$, the pursued algorithm seeks to find estimates of $\pi_{\textrm{opt}}$ along with the corresponding optimal maps $\{T_i\}$. 
In particular, let $(X^{(i)}_t)_{t\ge 0}: \Omega \to \mathcal{X}_i$ be a Markov process on the sample space $\Omega$, which is initialized by $\hat{X}^{(i)}$ and belongs to the Hilbert space $\mathcal{H}=\mathcal{L}^2(\Omega,\mathcal{F}_t, \mathbb{P}^{(i)})$. The latter is the space of square-integrable functions, which map $\Omega$ to $\mathcal{X}_i$ and are $\mathcal{F}_t$-measurable at time $t$, with the probability function $\mathbb{P}^{(i)}$. 
Our plan is to devise a  process such that $\hat{\pi}_t=1/N_p\sum_{i=1}^{N_p} \delta_{X_{i,t}^{(1)},...,X_{i,t}^{(K)}}$ approximates $\pi_{\textrm{opt}}$ as $t$ becomes large. As a by-product, the samples of the process will be regressed to recover the maps $\{T_i\}$. 
We denote by $\tilde{\pi}_t^{{X_i^{(s)}\leftrightarrow X_j^{(s)}}}$ the empirical joint measure $\tilde{\pi}_t$ updated by the sample swap between $X_i^{(s)}$ and $X_j^{(s)}$ of marginal $s$. 

\section{Main Idea}
\label{sec:main_idea}
\noindent Given independent and identically distributed (i.i.d.) samples $\{\hat{X}^{(i)}\}$, we develop a stochastic update rule that guides the resulting realizations toward approximating the optimal solution of \eqref{eq:mm-ot-gen}. Our objective is to construct this update rule in such a way that the computational complexity of each iteration scales linearly with the number of samples. A natural representation of the distribution based on given samples is via empirical measure
\begin{eqnarray}
\label{eq:disc-marg}
\tilde{\mu}_i= \frac{1}{N_p}\sum_{j=1}^{N_p}\delta_{\hat{X}^{(i)}_j} \ .
\end{eqnarray}
We leverage several key ideas in order to proceed. Let us focus on two marginal setup, i.e. $K=2$, and recall the following facts \cite{villani2009optimal,thorpe2018introduction,cuesta1997optimal}.
\begin{enumerate}
\item For discrete measures of the form \eqref{eq:disc-marg}, the optimal cost is given by
\begin{flalign}
\min_{\pi\in\Pi(\tilde{\mu}_1,\tilde{\mu}_2)} & \int c(x_1,x_2)\pi (dx)
=\min_{\gamma\in B^{N_p}}\frac{1}{N_p}\sum_{i,j}c(\hat{X}^{(1)}_i,\hat{X}^{(2)}_j)\gamma_{ij}\,
\end{flalign}
where $B^{N_p}$ is the set of $N_p\times N_p$ bistochastic matrices. 
\item The extremal points of $B^{N_p}$ are permutation matrices. Therefore
\begin{eqnarray}
\tilde{\pi}^{\textrm{opt}}=\min_{\sigma\in \Sigma^{N_p}}\frac{1}{N_p}\sum_{i=1}^{N_p}\delta_{(\hat{X}_i^{(1)},\hat{X}_{\sigma(i)}^{(2)})}
\end{eqnarray}
where $\Sigma^{N_p}$ is the set of permutations of $\{1,..,N_p\}$. The corresponding optimal map is given by $\tilde{T}(\hat{X}_i^{(1)})=\hat{X}^{(2)}_{\sigma^{\textrm{opt}} (i)}$.
\item For the $L^2$ cost, as $N_p\to\infty$, the optimal distribution and map weakly converge to the solution of the Monge-Kantorovich problem, i.e. $(\tilde{\pi}^{\textrm{opt}},\tilde{T})\rightharpoonup (\pi^{\textrm{opt}},T)$. 
\end{enumerate}
Hence the equivalent form of optimization problem \eqref{eq:mm-ot-gen}, admitting the mentioned assumptions, is given by a search over permutation matrices. In general, this remains a computationally intensive task, see e.g. the Hungarian algorithm \cite{kuhn1955hungarian}. To address this, a nested approach for finding a nearly optimal permutation matrix was proposed in \cite{puccetti2017algorithm,puccetti2020computation}. The Iterative Swapping Algorithm (ISA) aims to identify a near-optimal permutation by performing pairwise index swaps that reduce the cost. However, because ISA examines all possible swaps, its computational complexity remains quadratic. \\ \ \\
Starting from the premise that pairwise index swapping can yield near-optimal permutations, we introduce a stochastic variant of the ISA by drawing on analogies with Boltzmann kinetics. Rather than exhaustively examining all possible pairwise swaps, our approach involves randomly grouping indices, with each group containing only one swapping pair. Therefore at each iteration, only $N_p/2$ swapping candidates are assessed (instead of $N_p(N_p-1)/2$ required in ISA). This reduction simplifies the complexity of our stochastic version to linear scaling with respect to $N_p$.
\\ \ \\
Our scheme draws a close analogy to Boltzmann kinetics, and it is helpful to introduce the concept of particles to clarify this setup. Each particle represents a realization of a random variable, sampled from a marginal distribution. In this context, swapping can be viewed as a binary collision event. If accepted, the collision results in an index swap between two collision pairs. While a brute-force approach requires $N_p(N_p-1)/2$ collision pairs to be checked at every iteration, Bird \cite{Bird1963} introduced a randomization technique that requires only $N_p/2$ collision pairs to be considered, without introducing bias—as long as the selection of collision pairs is independent of the collision updates. This randomization concept has since been extended in fields such as stochastic gradient descent and mini-batch molecular dynamics.
Building on these insights, we review ISA and present collision-based dynamics, followed by a heuristic Boltzmann-like kinetic equation.

\section{Process Formulations}

\noindent  Consider discrete time index $t\in\{0,1,...\}$ and i.i.d. samples $\hat{X}^{(i)}_j$ for marginal $i$ and sample index $j=1,...,N_p$. 
\begin{enumerate}
\item {\it ISA process:}
For each marginal $i\in\{1,...,K\}$ and samples $j,k\in\{1,...,N_p\}$ with $k\ge j$, ISA updates the samples via
\begin{flalign}
(X^{(i)}_{j,t+1},X^{(i)}_{k,t+1})^T=
\mathcal{K}_{j,k}(X^{(i)}_{j,t},X^{(i)}_{k,t})^T.
\label{eq:swap}
\end{flalign}
The swaps are guided by the discrete cost
\begin{eqnarray}
m(\tilde{\pi}_t)=\mathbb{E}_{\tilde{\pi}_t}[c]
\label{eq:cost}
\end{eqnarray}
where $\tilde{\pi}_t$ is the empirical measure of $X_t$. The swapping kernel is given by 
\begin{flalign}
\mathcal{K}_{j,k}=\begin{cases}
 {I}_{2n\times2n} & \text{if} \ m(\tilde{\pi}_t^{{X_j^{(i)}\leftrightarrow X_k^{(i)}}})\ge m(\tilde{\pi}_t) \\
 {J}_{2n\times2n} &  \text{if} \  m(\tilde{\pi}_t^{{X_j^{(i)}\leftrightarrow X_k^{(i)}}})< m(\tilde{\pi}_t)
 \end{cases}
 \label{eq:swap_kernel}
\end{flalign}

with $I_{n \times n}$ as the identity matrix and $J$ an exchange matrix of the form
\begin{eqnarray}
{J}_{2n\times 2n}&=&\begin{bmatrix}
{0}_{n\times n} & {I}_{n\times n} \\
{I}_{n\times n} & {0}_{n\times n} 
\end{bmatrix}
\end{eqnarray}
and ${0}_{n\times n}$ is a $n \times n$ matrix with zero entries.  The swapping kernel \eqref{eq:swap_kernel} allows swaps if it leads to reduction in the cost associated with the empirical measure \eqref{eq:cost}. 
\item {\it Collision-based dynamics:} The proposed collision-based version of ISA performs similar steps with the difference that $j,k$ are now chosen from a random subset $\mathcal{C}\subset \{1,...,N_p\}$ of size 2. Therefore, in the collision-based method, instead of applying \eqref{eq:swap} to all pairs $k,j\in\{1,...,N_p\}$ with $k\ge j$, we pick $j,k\sim \mathcal{U}([1,N_p])$ where $\mathcal{U}(.)$ is a discrete uniform measure with values between $1$ and $N_p$. As a result, the complexity of the algorithm reduces to $\mathcal{O}(N_p)$. Note that this randomization step in general can be justified as long as the subsets are sampled independent of the random variable $X$. While the consistency proofs exist for range of kernels \cite{liu2024random}, we leave the theoretical 
consistency between collision-based process and ISA to separate studies.
\item {\it Boltzmann kinetics:}  There is a close analogy between the proposed collision process and Boltzmann kinetics. To simplify the illustration, let us consider a setup of two marginals $\{\mu_1,\mu_2\}$. The proposed collision process evolves an initial joint measure of $\{\mu_1,\mu_2\}$ in a fashion similar to binary collisions, where collisions refer to swapping the state of two particles. Given that the collision here simply exchanges the sample values, the equivalent Boltzmann operator takes a concise form. Let $\rho_t$ be the time dependent density of the joint measure.
An equivalent collision operator of the Boltzmann-type can be described as
\begin{flalign}
Q[\rho_t,\rho_t]&=
\int_{\mathbb{R}^{2n}}\rho_t(x_1,y)\rho_t(x,y_1)\Omega(x,x_1,y,y_1) dx_1dy_1-\alpha(x,y) \rho_t(x,y),
\label{eq:BoltzmannCollOperator}\\
\alpha(x,y)&=
\int_{\mathbb{R}^{2n}}\rho_t(x_1,y_1)\Omega(x,x_1,y,y_1) dx_1dy_1
\label{eq:alpha}
\end{flalign}
and the collision kernel reads
\begin{flalign}
\Omega(x,x_1,y,y_1)=H\bigg(&c(x,y)+c(x_1,y_1)
-c(x_1,y)-c(x,y_1)\bigg) 
\end{flalign}
and $H(.)$ is the Heaviside function. Heuristically, the kinetic model \eqref{eq:BoltzmannCollOperator} describes a process where binary collisions are only accepted if the cost $c$ is decreased by the swaps between the two randomly picked sample points. However, one should note that the Boltzmann kinetics operate on the continuous time whereas the proposed collision process is discrete in time. Although consistency between the two descriptions may be proven following the recipe provided by Wagner's proof of Monte Carlo solution to the Boltzmann equation \cite{wagner1992convergence}, we leave out the theoretical justifications for future works.
\end{enumerate}

\section{Monte Carlo Solution Algorithm for the Collisional dynamics}
\label{sec:MonteCarloSolution}
\noindent Motivated by the direct Monte Carlo solution algorithms to the Boltzmann \cite{Bird} and the Fokker-Planck equation \cite{takizuka1977binary} for rarefied gas and plasma dynamics, here we devise a collision-based numerical scheme to solve the discrete optimal transport problem. In order to ensure that all the particles are considered for the collision in one time step, we consider the following collision routine for each marginal:
\begin{itemize}
    \item[-] Generate a random list of particle indices $R$ of size $N_p$ without repetition.
    \item[-] Decompose $R$ into two subsets of the same size $I$ and $J$ where $I  \cap J = \varnothing$.
    \item[-] Swap particles with indices $I_k$ and $J_k$ for collisions using \eqref{eq:swap} where $k=1,...,N_p/2$.
\end{itemize}
We note that by shuffling the particle indices, one can easily find a random list of particle indices $R$. In Algorithm~\ref{alg:collOT}, we give a detailed description of the proposed method. 
\begin{algorithm}
 \caption{Collision-based algorithm to MMOT problem}
   \label{alg:collOT}
   \begin{algorithmic}
\STATE{\bfseries Input:} $X:=[ X^{(1)}, ... ,  X^{(K)} ]$ and tolerance $\hat \epsilon$\;
\REPEAT
 \FOR{$i=1,\hdots,K$}
    \STATE Generate an even random list of particle indices $R$.\;
    \STATE Decompose $R$ into same-size subsets $I$ and $J$ where $I  \cap J = \varnothing$ and $|I|=|J|=\lfloor N_p/2 \rfloor$.\;
    \FOR{$k=1,\hdots,\lfloor N_p/2 \rfloor$}
    \IF{$m(\hat{\pi}_t^{{X_{I_k}^{(i)}\leftrightarrow X_{J_k}^{(i)}}})< m(\hat{\pi}_t)$}
        \STATE $X^{(i)}_{I_k} \leftarrow X^{(i)}_{J_k}$ and 
        $X^{(i)}_{J_k} \leftarrow X^{(i)}_{I_k}$.\;
        \ENDIF
    \ENDFOR
 \ENDFOR
 \UNTIL{Convergence in $\mathbb{E}_{\hat \pi_t}[ c(X_t^{(1)},...,X_t^{(K)}) ]$ with tolerance $\hat \epsilon$}
 \STATE {\bfseries Output:} $X$
 \end{algorithmic}
\end{algorithm}

\section{Properties of Collisional Dynamics for the Optimal Transport Problem}
\label{sec:properties_method}

\noindent The proposed collision-based Monte Carlo solution Algorithm~\ref{alg:collOT} has several numerical properties that we list next.
\begin{itemize}
    \item \noindent \textbf{Marginal preservation.} Since we only change the order of particles in each marginal when a collision is accepted, Algorithm~\ref{alg:collOT} preserves the marginals up to machine accuracy on the discrete points.  

\item \noindent \textbf{Monotone convergence.} 
Collisions are only accepted if they reduce the cost of the optimal transport problem. This guarantees that Algorithm~\ref{alg:collOT} converges to the stationary solution monotonically. However, finding the convergence rate is not trivial given the discontinuity of the jump process. If the proposed collision-based dynamics behaves similar to the Boltzmann kinetics, we expect that the proposed Algorithm~\ref{alg:collOT} converges exponentially to its stationary solution \cite{desvillettes2010celebrating}. In particular, we expect that $\hat \pi$ exponentially converges to stationary $\hat \pi_\mathrm{st}$, i.e. the error $\epsilon := |(\hat \pi(t)-\hat \pi_\mathrm{st})/(\hat \pi(0)-\hat \pi_\mathrm{st})|$ follows 
 \begin{flalign}
        \epsilon = \mathcal{O}(e^{-\hat \alpha t})~
        \label{eq:exp_relax}
    \end{flalign} 
where $\hat \alpha$ denotes the upper bound of $\alpha$ defined in eq.~\eqref{eq:alpha}. To see the exponential convergence, let us consider two marginal OT problem. Following  \cite{wild1951boltzmann}, \cite{carlen2000central}, \cite{gabetta1997relaxation} and  \cite{pareschi2005numerical}, let us consider the Cauchy problem
\begin{flalign}
    \frac{\partial \rho}{\partial t} = P[\rho, \rho]- \hat \alpha \rho
\end{flalign}
where $P[\rho,\rho]$ is a bilinear operator, and $\hat \alpha\neq 0$ is a constant.
The solution to the Cauchy problem can be written as
\begin{flalign}
    \rho = e^{-\hat \alpha t} \sum_{k=0}^{\infty} (1-e^{-\hat \alpha t})^k \rho_k
    \label{eq:sol_cauchy}
\end{flalign}
where $\rho_k$ is given by the recurrence formula
\begin{flalign}
\rho_k=\frac{1}{k+1} \sum_{h=0}^k \frac{1}{\hat \alpha}P[\rho_h,\rho_{k-h}].
\end{flalign}
By defining $P[\rho,\rho]:=Q[\rho,\rho] + \hat \alpha \rho$, formally we have
 $ \lim_{k\rightarrow \infty} \rho_k = \lim_{t\rightarrow \infty} \rho  = \rho^*$,
where $\rho^*$ is the equilibrium solution to the Boltzmann equation, i.e. the target sub-optimal joint density in this context. The solution to the Cauchy problem Eq.~\eqref{eq:sol_cauchy}, which is used to solve the Boltzmann equation, admits the exponential convergence to the stationary solution \eqref{eq:exp_relax} for a fixed $\hat \alpha$. See Appendix~\ref{sec:wild_expansion} for more details.
However, the stationary solution may not be optimal as it only ensures that no further improvement is possible through binary swapping among collision pairs.
In other words, the proposed algorithm converges exponentially to a near-optimal solution $\hat \pi_\mathrm{st}\approx \hat{\pi}_{\textrm{opt}}$, as long as $\alpha$ remains finite. In the numerical tests presented in \ref{sec:results}, we show in several examples that the recovered near-optimal solution has a reasonably small relative error compared to EMD, making it useful for practical purposes.


\item \noindent \textbf{Affordable computational complexity.} Each iteration of Algorithm~\ref{alg:collOT} for a given marginal has the computational complexity of $\mathcal{O}(\beta N_p)$ where $\beta$ denotes the cost of computing collision probability for a collision candidate and $N_p$ is number of particles per marginal.  In the case of the optimal map corresponding to the $L^p$-Wasserstein distance, we have $\beta=\mathcal{O}(nK)$. This becomes possible since the collision probability between $i$th and $j$th particle in the $k$th marginal is computed using the change in the cost, i.e.,
    \begin{flalign}
     \sum_{l=1, l\neq k}^{K} \lVert {X_i^{(l)}} - {X_j^{(k)}}\rVert_p^p + \lVert {X_j^{(l)}} 
     - {X_i^{(k)}}\rVert_p^p
      - \lVert {X_i^{(l)}} - {X_i^{(k)}}\rVert_p^p - \lVert {X_j^{(l)}} - {X_j^{(k)}}\rVert_p^p~. \nonumber
    \end{flalign}
    This also implies that the computational complexity with respect to the number of marginals $K$ is $\mathcal{O}(K)$. Overall, we expect Algorithm~\ref{alg:collOT} to have computational complexity of $\mathcal{O}(n K^2 N_p)$ for $L^p$-Wasserstein distance, given $K$ marginals, $N_p$ samples per marginal, and $n$-dimensional sample space. 
    
\item \noindent \textbf{Low memory consumption.} Since the proposed method does not require computing any distance matrix at any point, which is often used in EMD and Sinkhorn method, it has a more affordable memory consumption of $\mathcal{O}(n K N_p)$ which is of the same order as the input.

\item  \noindent \textbf{Relaxed constraints on the cost function.} Our scheme, unlike gradient based methods, does not require regularity conditions on the cost function $c(.)$. In other words, the algorithmic steps of the proposed collision-based dynamics can be performed irrespective of the regularity of $c(.)$.  We expect that the proposed method leads to accurate results as long as the original OT problem is well-posed. 

\item  \noindent \textbf{Constant weight.} In the proposed method, we assumed that the weight of all particles are equal and remain constant in OT problem. This implies that if the input samples are weighted, a resampling method needs to be used to enforce constant weight for all samples. While we see this as a limitation, we believe the proposed numerical scheme may be adapted by the stochastic weighted particle method \cite{rjasanow1996stochastic} to allow varying weight to bypass resampling.

\item  \noindent \textbf{No data race for collisions in each marginal.} In the collision step for each marginal, the Algorithm~\ref{alg:collOT} tests unique pairs of particles for collisions by construction. Therefore, collisions in each marginals can be trivially parallelized since there is no data race.
\end{itemize}

\section{Results}
\label{sec:results}

\noindent Here, we test the proposed collision-based Algorithm~\ref{alg:collOT} in solving MMOT as a metric to find distances between images in a dataset. In Appendix~\ref{sec:further_results}, we carry out further test on several toy problems to validate the convergence rate and computational cost of the proposed method compared to EMD and Sinkhorn. Everywhere in this study, we report an estimate of Wasserstein distance $d^p(.)$ given samples of $i$the marginal  $X^{(i)}\in \mathbb{R}^{ N_p \times n}$ for $i=1,...,K$, i.e. $d^p(X):=
\sum_{j>k}^{K} \sum_{i=1}^{N_p} \lVert X_i^{(j)}-X_i^{(k)}\rVert_p^p/N_p$. All computations are done on a laptop with an Intel Core i7-8550U CPU that runs with $1.8$GHz frequency equipped with $16$ GB memory. In this paper we use Python Optimal Transport library \cite{flamary2021pot} for EMD and Sinkhorn computations.


\noindent One of the applications of Wasserstein distance is labeling datasets, since it provides us with a metric in the space of distributions. Here we show the efficiency of the proposed collision-based solution to MMOT by treating this problem as one. Consider the Japanese Female Facial Expression (JAFFE) \cite{lyons1998japanese},  butterfly \cite{chen2018fine}, and CelebA datasets \cite{liu2015faceattributes}. The JAFFE dataset consists of $213$ images, where we treat each as a marginal. From  the butterfly dataset, we consider $50$ classes, from which we select $4$ pictures randomly which leads to MMOT problem with $200$ marginals. Similarly, we randomly select $200$ images  from the CelebA dataset for the MMOT problem. 

\begin{figure}
    \centering 
    \begin{tabular}{ccc}
    \hspace{-2cm} 
\includegraphics[width=0.37\linewidth]{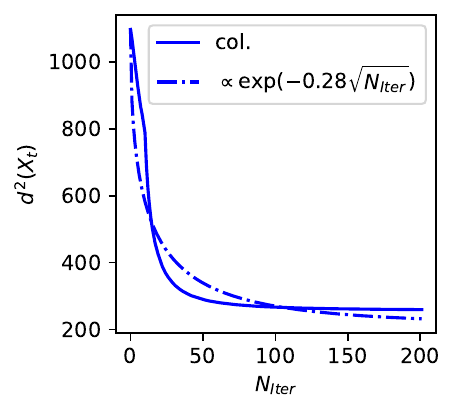}
&
\hspace{-2cm} 
\includegraphics[width=0.37\linewidth]{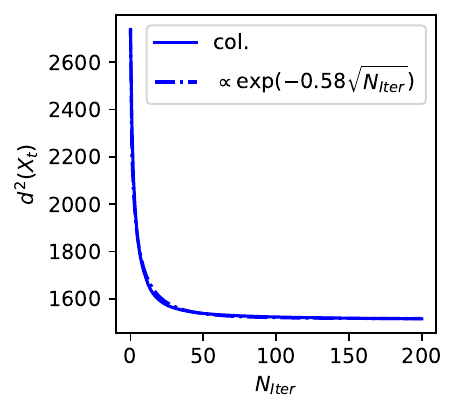}
&
\hspace{-2cm} 
\includegraphics[width=0.37\linewidth]{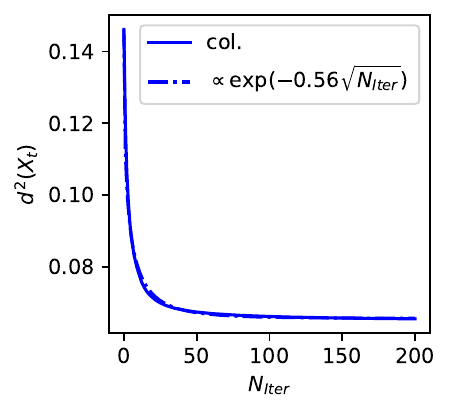}
\\
\hspace{-1cm} (a) & \hspace{-1cm}  (b) &  \hspace{-1cm}  (c)
\\
\hspace{-2cm} 
\includegraphics[width=0.39\linewidth]{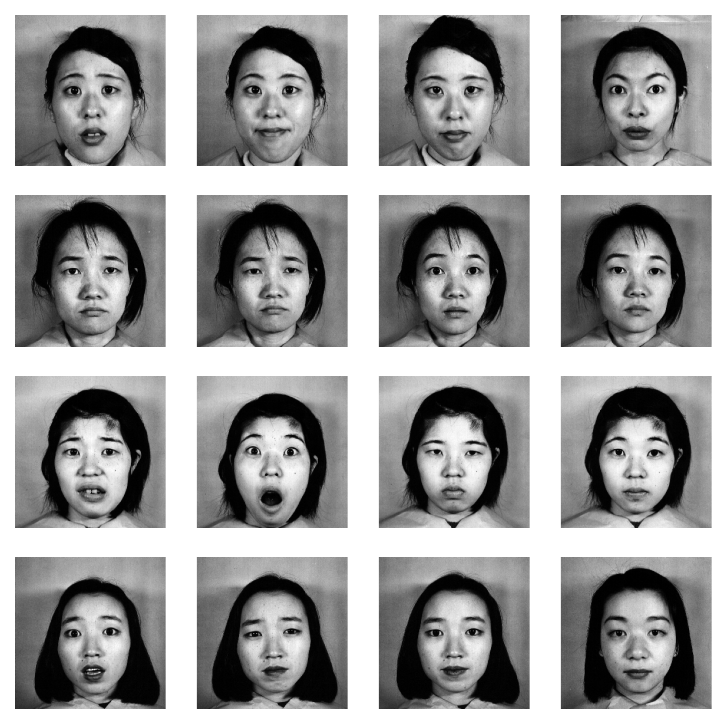}
& 
\includegraphics[width=0.4\linewidth]{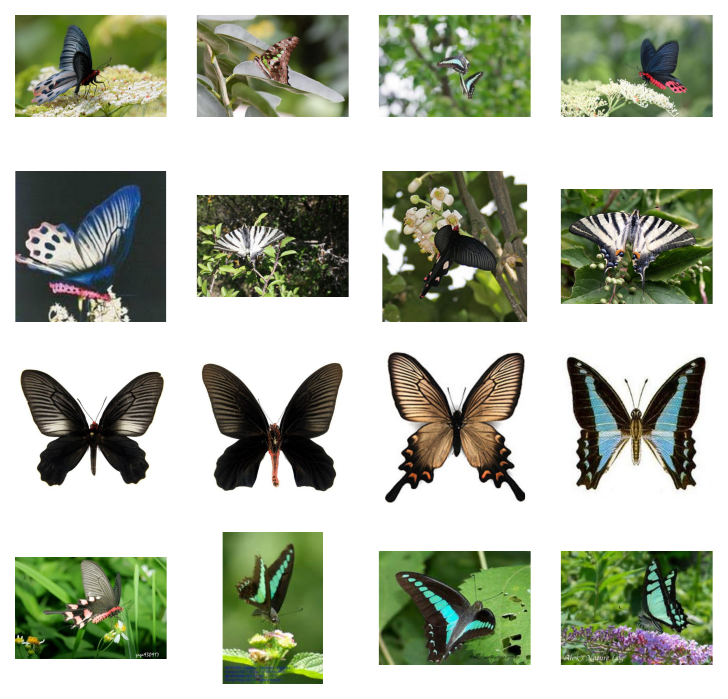}
& 
\includegraphics[width=0.33\linewidth]{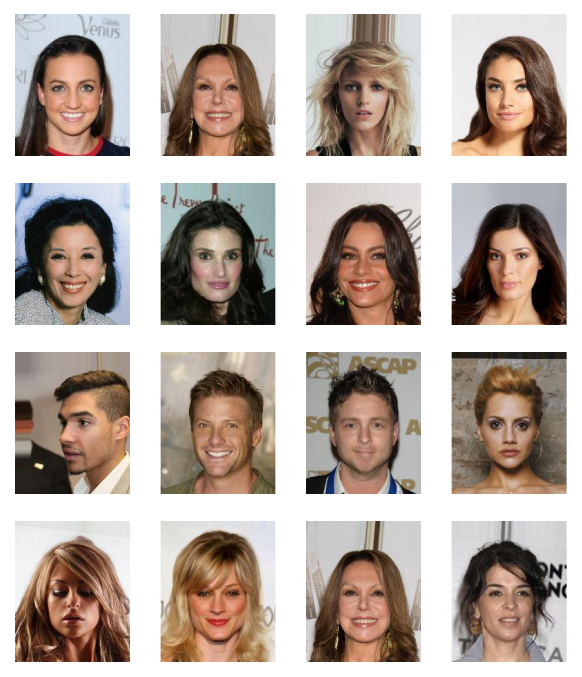}
\\ 
\hspace{-2cm} (d) & (e) & (f)
    \end{tabular}
\caption{Evolution of Wasserstein distance against number of iteration $N_\mathrm{Iter}$ for collision based OT algorithm \ref{alg:collOT} (a)-(c) for JAFFE, Butterfly, and CelebA dataset, as well as drawing 4 closest pictures from each dataset for 4 random samples using the found distribution of Wasserstein distance in the dataset (d)-(f). }
    \label{fig:japanese}
\end{figure}

\begin{figure}
    \centering 
    \begin{tabular}{cccc}
    \hspace{-2cm}
\includegraphics[width=0.32\linewidth]{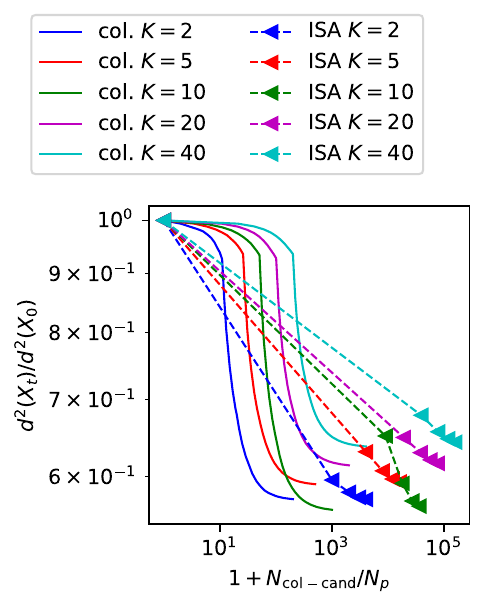}
& 
\includegraphics[width=0.3\linewidth]{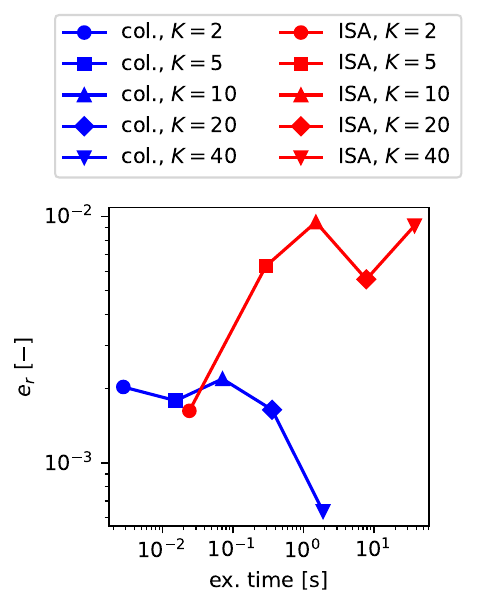}
& 
\hspace{-0.6cm}
\includegraphics[width=0.3\linewidth]{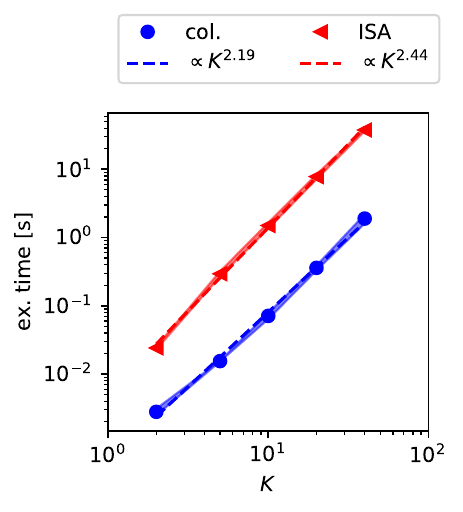}
    &
    \hspace{-0.6cm}
    \includegraphics[width=0.31\linewidth]{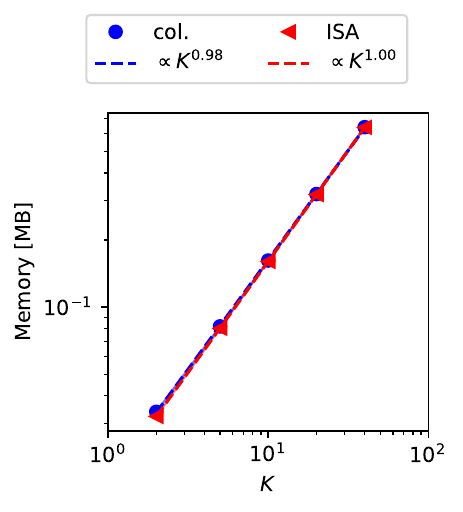}
    \\
    \hspace{-0.6cm} (a) & (b) & (c) & (d)
    \end{tabular}
    \caption{Evolution of Wasserstein distance estimate as a function of number of collision candidates $N_\mathrm{col-cand}$ per number of samples $N_p$ (a), relative error against execution time (b), scaling of execution time (c), and memory footprint (d) versus the number of considered marginals $K$ for the multi-marginal optimal map problem in JAFFE dataset. In order to have similar orders of magnitude in the relative error, we consider collisional OT with 200 iterations and ISA with 4 iterations. Here, the
solution obtained with ISA using 10 iterations is considered as the reference solution to compute the relative error.}
    \label{fig:japanese_w2_cost_Nm}
\end{figure}

\begin{figure}[htbp!]
    \centering 
    \begin{tabular}{cccc}
\hspace{-2cm}    \includegraphics[width=0.3\linewidth]{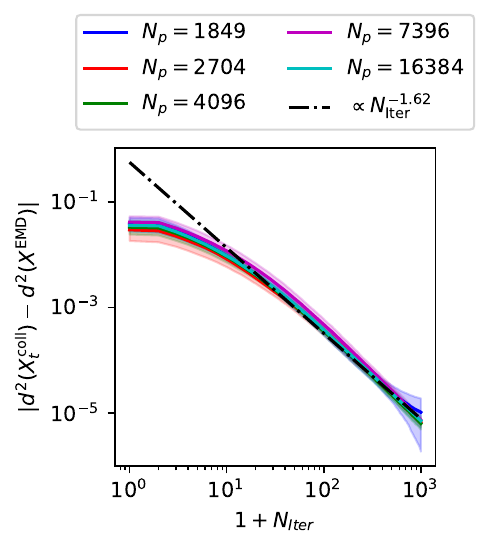}
    &
    \hspace{-0.6cm}
    \includegraphics[width=0.3\linewidth]{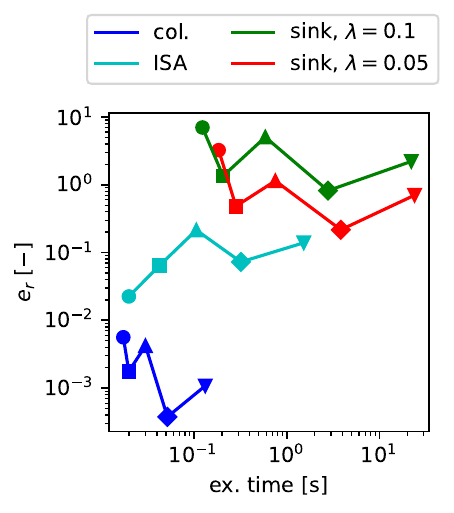}
    &
    \hspace{-0.6cm}
\includegraphics[width=0.3\linewidth]{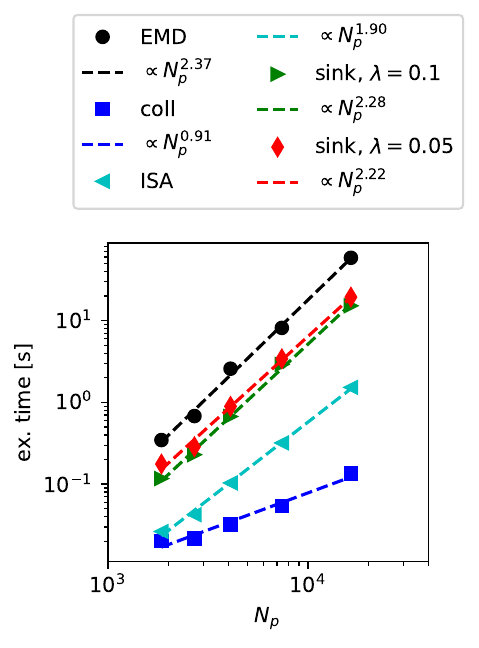}
    
    &
    \hspace{-0.6cm}
    \includegraphics[width=0.3\linewidth]{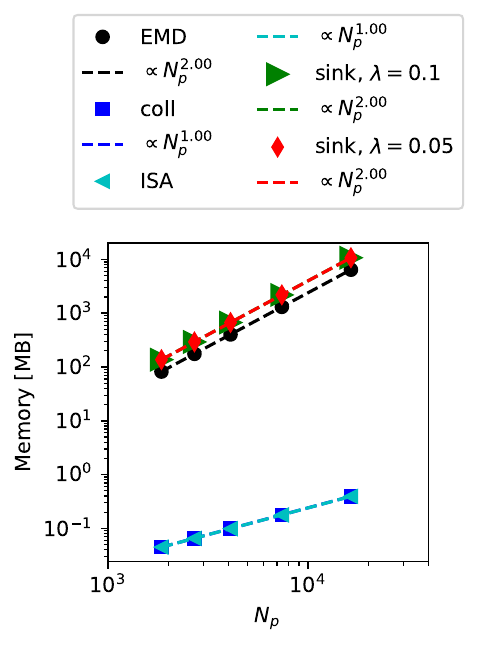}
    \\
    \hspace{-1.5cm} (a) & (b) & (c) & (d)
    \end{tabular}
    \caption{Optimal transport between 10 randomly selected pairs of pictures from JAFFE dataset, each for a range of the number of pixels denoted by $N_p$. Here we investigate the solution from collision-based Algorithm~\ref{alg:collOT} by plotting error from EMD (a), relative error against execution time (b), scaling of  execution time (c), and memory footprint (d) compared to EMD and Sinkhorn with a variation of regularization factor $\lambda=0.1$ and $0.05$. In (b), the symbols $\circ$, $\square$, $\bigtriangleup$, $\Diamond$, $\triangledown$ correspond to $N_p=43^2,52^2,64^2,86^2,128^2$, respectively.}
    \label{fig:japanese_butterfly_error_cost}
\end{figure}

We deploy the collision-based solution Algorithm~\ref{alg:collOT} to these MMOT problems with $L^2$-Wasserstein cost to find the pairs of particles/samples across marginals. Since we are minimizing the total cost $\sum_{j>k}^{K} \sum_{i=1}^{N_p} \lVert X_i^{(j)}-X_i^{(k)}\rVert_2^2/N_p$, we are also approximating the optimal map between every two $j,k$ marginals that minimizes $\sum_{i=1}^{N_p} \lVert X_i^{(j)}-X_i^{(k)}\rVert_2^2/N_p$. As shown in Fig. \ref{fig:japanese}-\ref{fig:japanese_w2_cost_Nm}, the proposed collision MMOT can find the pair-wised Wasserstein distance distribution in both datasets efficiently. We also observe that the convergence rate is not affected by the number of marginals. As expected, the execution time scales $\mathcal{O}(K^2)$ and memory $\mathcal{O}(K)$.
\\ \ \\
\noindent Here, we also compare the solution to optimal transport between two randomly selected pictures (marginals) from JAFFE dataset with the benchmark. As shown in Fig.~\ref{fig:japanese_butterfly_error_cost}, we observe convergence to the EMD solution with an error of  $\epsilon \propto N_\mathrm{Iter}^{-1.6 }$
. Furthermore, we validate the computational complexity reported in section~\ref{sec:properties_method}, and observe that the proposed collisional OT method outperforms EMD and Sinkhorn with respect to execution time and memory consumption. For further numerical tests, see Appendix~\ref{sec:app_wasserstein_dataset}.

\section{Conclusion}
\label{sec:conclusion}
\noindent We proposed a novel solution algorithm called collision-based dynamics for the discrete OT problem, including multi-marginal settings. The devised collision process is based on random binary swaps of the samples and is built on close analogy with the Boltzmann kinetics. We showed that in the case of $L^p$-Wasserstein distance, the proposed method has the computational complexity of $\mathcal{O}(n K^2 N_p)$, where $n$ is the dimension of each sample, $N_p$ is the number particles/samples per marginal, and $K$ is the number of marginals. We achieved this performance by randomizing the swapping process. The method conserves marginals by construction. We showed empirically that it admits an exponential convergence to a near-optimal solution.
\\ \ \\
We investigated the computational cost, optimality gap, and memory consumption of the collision process in several toy problems, and validated our estimates on the cost and memory requirements. Furthermore, we showed the capability of the proposed method in finding the optimal map in a five-marginal setting. Moreover, we tested the algorithm to find an optimal map between pictures in a dataset, treating it as a multi-marginal OT problem. The proposed collision-based dynamics proves to be highly efficient, e.g., in comparison to the Sinkhorn algorithm. We anticipate broad applications of the devised method in various settings where multi-marginal OT problem is of relevance, including Density Functional Theory \cite{buttazzo2012optimal}, among others. 

 \bibliographystyle{plainnat}
  \bibliography{refs}

\newpage
\appendix
\onecolumn

\section{Wild expansion for the Bolztmann equation}
\label{sec:wild_expansion}

In this section, we review the Wild expansion for the Boltzmann equation which is used as the basis to justify the exponential behavior of the collisional OT. Let us revisit the Botlzmann equation. The particle-particle interaction is modeled via a collision operator
\begin{flalign}
Q[\rho_t,\rho_t]=
 P[\rho_t,\rho_t] -\alpha_t \rho_t
\end{flalign}
where
\begin{flalign}
P[ \rho_t (x,y), \rho_t (x,y)] &= \int_{\mathbb{R}^{2n}}\rho_t(x_1,y)\rho_t(x,y_1)\Omega(x,x_1,y,y_1) dx_1dy_1
\end{flalign}
and
\begin{flalign}
\alpha_t(x,y)&=
\int_{\mathbb{R}^{2n}}\rho_t(x_1,y_1)\Omega(x,x_1,y,y_1) dx_1dy_1~.
\end{flalign}
Then, the Boltzmann equation as a kinetic integro-differential model can be written as
\begin{flalign}
    \frac{\partial \rho_t}{\partial t} = P[\rho_t,  \rho_t] - \alpha_t \rho_t~.
    \label{eq:boltz}
\end{flalign}


Assuming constant $\alpha_t\equiv \hat{\alpha}$, multiplying both sides by $\exp(\int \hat{\alpha} d \tau^\prime)$,  and integrating in the time span $[t_0,t]$, we get
\begin{flalign}
    \rho_t = \rho_{t_0}  \exp \left(- \hat \alpha  t \right) + \int_{t_0}^t \exp\left(- \hat \alpha  (t-\tau) \right) P[\rho_\tau, \rho_\tau] d \tau \ . 
\end{flalign}
The solution to this equation can be obtained using backward particle tracking method \cite{wild1951boltzmann}. Consider the notation
\begin{flalign}
    \rho_{0} = \rho_{t_0} \exp \left(- \hat \alpha  t \right)
\end{flalign}
and for any function $G_t(x,y)$, define
\begin{flalign}
    S\{ G \} := \int_{t_0}^t \exp \left(- \hat \alpha (t-\tau) \right)  G_\tau(x,y) d\tau \ .
\end{flalign}
Therefore we have
\begin{flalign}
    \rho = \rho_{0} + S\{ P[\rho , \rho ] \}~.
\end{flalign}

Assuming existence of $\rho$ and substituting $\rho$ in the last term by itself lead to

\begin{flalign}
    \rho_{r+1}=\rho_0 +S \{ P[\rho_r, \rho_r] \}~.
\end{flalign}
Since $\rho$ is non-negative, we have
\begin{flalign}
    0 \leq \rho_r \leq \rho_s \leq \rho \ \ \ 
    \textrm{for\ all}\ \ r \leq s~.
\end{flalign}
The functions $\rho_r$ constitute an increasing sequence which is bounded above, and thus convergent with the limiting function satisfying the integral  equation

\begin{flalign}
    \rho_t(x,y) = \rho_{t_0}(x,y) e^{-\hat \alpha t} + \int_{t_0}^t e^{-\hat \alpha(t-\tau)} P[\rho_t, \rho_t]  d\tau~.
\end{flalign}

In order to find how fast the solution converges, let us consider the complete partition of $n$ defined in \cite{wild1951boltzmann} denoted by $P_r(n)$ where
\begin{flalign}
    P_r(n) = P_s(m) P_t(n-m)~.
\end{flalign}
By  induction, we have

\begin{flalign}
    F^{P(1)}&= \rho_{t_0},\\
    F^{P_r(n)} &= P[F^{P_s(m)} , F^{P_t(n-m)}] \ .
\end{flalign}
Furthermore, consider a numerical function of $P_r(n)$ as $g_r(n)$ by the relations
\begin{flalign}
    g(1) &= 1
    \\
    g_r(n) &= \frac{1}{n-1} g_s(m) g_t(n-m)
\end{flalign}
where $\sum_r g_r=1$ for all $n$. This leads to
\begin{flalign}
    \rho_t(x,y) = e^{-\hat \alpha t} \sum_{n=1}^\infty (1-e^{-\hat \alpha t})^{n-1} \sum_r g_r(n) F^{P_r(n)}.
\end{flalign}

This solution can be easily verified by substitution 

\begin{flalign}
    \int_0^t &e^{-\hat \alpha (t-\tau)} \alpha P[\rho_\tau, \rho_\tau]  d \tau \nonumber \\
    & = e^{-\hat \alpha t} \int_0^t  e^{-\hat \alpha \tau} \sum_{n=1} \sum_{m=1} \left(  (1-e^{-\alpha r})^{n+m-2} \sum_{s,r} g_s(n) g_r(m) P[F^{P_s(n)}, F^{P_r(m)}]  \right) d \tau \nonumber \\
    & = e^{-\hat \alpha t} \sum_{n=2}^\infty \int_0^t (n-1)   e^{-\hat \alpha \tau}  (1-e^{-\hat \alpha r})^{n-2} d \tau \sum_{m=1}^{n-1} \frac{1}{n-1} \sum_{s,r} g_s(n) g_r(n-m) P[F^{P_s(n)}, F^{P_r(n-m)}] 
    \nonumber \\
    & = e^{-\hat \alpha t} \sum_{n=2}^\infty (1-e^{-\hat \alpha t})^{n-1} \sum_r g_r(n) F^{P_r(n)} \nonumber
    \\
    &= \rho_t - \rho_{t_0} e^{-\hat \alpha t}.
\end{flalign}

Consider $\rho^*$ as the equilibrium solution to the Boltzmann equation. As shown in \cite{wild1951boltzmann}, for a given $\epsilon$ and $t>t_0$, there exists constants $n_0$ and $K$ where $F^{P_r(n)}(x)<K$, such that
\begin{flalign}
    |\rho - \rho^*| < K n_0 e^{-\hat \alpha t_0} + \frac{2}{3} \epsilon e^{-\hat \alpha t} \sum_{n=1}^\infty (1-e^{-\hat \alpha t})^{n-1}~.
\end{flalign}

\section{Iterated Swapping Algorithm}

\noindent Here, we give a short description of ISA algorithm used in this paper, given its similarity to the collision-based method proposed in this paper. As detailed in Algorithm~\ref{alg:ISA_OT}, in each iteration for each marginal, $\mathcal{O}(N_p^2)$ operations are carried out, while each particle maybe be accepted for swap (collision) more than once. This makes the algorithm prone to data race as an issue for shared-memory parallelism. 

\begin{algorithm}
 \caption{Iterated Swapping Algorithm for the MMOT problem}
   \label{alg:ISA_OT}
\begin{algorithmic}

\STATE{\textbf Inputs:} $X := [ X^{(1)}, ... ,  X^{(K)} ]$ and tolerance $\hat \epsilon$.\;
\REPEAT
 \FOR{$i=1,\hdots,K$}
 \FOR{$j=1,\hdots,N_p$}
    \FOR{$k=j+1,\hdots, N_p$}
        \IF{$\sigma( X^{(i)}_{j},  X^{(i)}_{k};  X^{(i)}_{k}, X^{(i)}_{j} )=1$}
\STATE{
        $X^{(i)}_{j} \leftarrow X^{(i)}_{k}$ and 
        $X^{(i)}_{k} \leftarrow X^{(i)}_{j}$.\;}
        \ENDIF
    \ENDFOR
  \ENDFOR
 \ENDFOR
 \UNTIL{Convergence in $\mathbb{E}_{\hat \pi_t}[ c(X_t) ]$ with tolerance $\hat \epsilon$.}
 \STATE{\textbf{Ouput:}} $X$.\;
 \end{algorithmic}
\end{algorithm}

\section{Further Results}
\label{sec:further_results}

\noindent In this section, we test the convergence rate of the proposed collision-based solution algorithm to the optimal transport problem in several toy problems. The summary of results in terms of relative error versus computational time is shown in Table~\ref{table:overal_results}. Everywhere in this study, we report results with the measured uncertainty that is indicated with $\pm$ standard deviation.

\begin{table}[H]
{\footnotesize \hspace{-2.5cm}
\begin{tabular}{|c|c|c|c|c|c|c|}
\hline
Problem & & EMD & Sinkhorn ($\lambda_1$) & Sinkhorn ($\lambda_2$) & ISA & Collisional OT
\\
\hline
JAFFE & Rel. error &
-&
6.413e-1 ± 0.616&
1.796e-1 ± 0.122&
1.409e-2 ± 1.390e-2&
8.456e-3 ± 1.541e-2
\\
 & Time [s] & 9.134 ± 0.820&
2.659 ± 0.172&
3.665 ± 0.581&
0.470 ± 0.034&
\textbf{0.022 ± 0.003}
\\ \hline
Butterfly & Rel. error &
-& 2.824e-1 ± 1.652e-1&
2.717e-1 ± 8.056e-2&
9.796e-3 ± 2.636e-2&
1.028e-2 ± 2.860e-2
\\
 & Time [s] & 9.761 ±0.681&
2.765 ± 2.122e-1&
4.349 ± 1.713&
0.217 ± 2.190e-2&
\textbf{3.494e-2 ± 7.219e-3}
\\ \hline
CelebA & Rel. error &
-& 2.220 ± 2.897e-1 &
2.219 ± 2.896e-1 &
4.405e-3 ± 3.455e-3 &
4.643e-3 ± 1.926e-3
\\
 & Time [s] & 12.717 ± 0.864 &
2.479 ± 0.285 &
2.571 ± 0.346 &
0.212 ± 0.014 &
\textbf{0.120 ± 0.014}
\\ \hline
Swiss Roll-Normal & Rel. error &
-& 0.09 ± 0.01&
0.03 ± 0.01&
1.731e-2 ± 1.415e-3&
1.665e-2 ± 5.533e-4
\\
 & Time [s] & 14.235 ± 0.552&
4.09 ± 0.73&
5.30 ± 0.59&
0.130 ± 0.001&
\textbf{0.098 ± 0.004}
\\ \hline
Banana-Normal & Rel. error &
-& 0.29 ± 0.001&
0.11 ± 0.01&
9.164e-3 ± 3.611e-4&
1.239e-2 ± 7.734e-4
\\
 & Time [s] & 13.730 ± 0.148&
4.70 ± 0.06&
6.81 ± 0.05&
0.146 ± 0.021&
\textbf{0.114 ± 0.010}
\\ \hline
Funnel-Normal & Rel. error &
-& 0.43 ± 0.01&
0.18 ± 0.01&
1.497e-2 ± 7.107e-4&
1.550e-2 ± 7.087e-4
\\
 & Time [s] & 13.489 ± 0.431&
4.65 ± 0.43&
6.77 ± 0.56&
0.155 ± 0.025&
\textbf{0.105 ± 0.001}
\\ \hline
Ring-Normal & Rel. error &
-& 0.32 ± 0.01&
0.29 ± 0.01&
1.715e-2 ± 9.159e-4&
1.735e-2 ± 5.948e-4
\\
 & Time [s] & 10.391 ± 0.552&
2.80 ± 0.15&
3.43 ± 0.25&
0.167 ± 0.001&
\textbf{0.119 ± 0.008}
\\ \hline
\end{tabular}
}
\caption{Execution time and relative error of Sinkhron with two regularization factors $\lambda_1>\lambda_2$, ISA, and collisional method for the considered test cases, each obtained using 8000 samples and repeated 20 times to obtain reasonable statistics. Here, we stop ISA and Collisional OT algorithms with a similar tolerance of convergence.}
\label{table:overal_results}
\end{table}

\subsection{Learning a five-marginal map}
\label{sec:5marginal}

\begin{figure*}[t]
    \centering
    \begin{tikzpicture}
        \node (note0) at (-8, 0) {\rotatebox{90}{Input}};
        
        \node (top) at (-6, 0) {\includegraphics[width=0.25\textwidth]{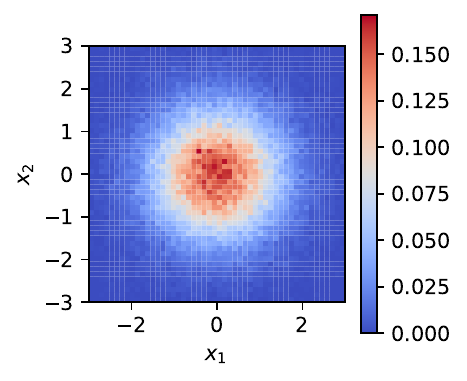}};

        \node (note1) at (-14, -4) {\rotatebox{90}{Output}};
        
        \node (fig1) at (-12, -4) {\includegraphics[width=0.25\textwidth]{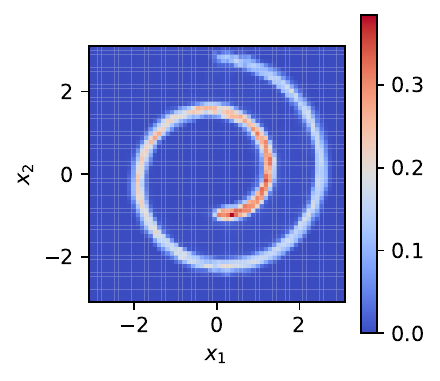}};
        \node (fig2) at (-8, -4) {\includegraphics[width=0.25\textwidth]{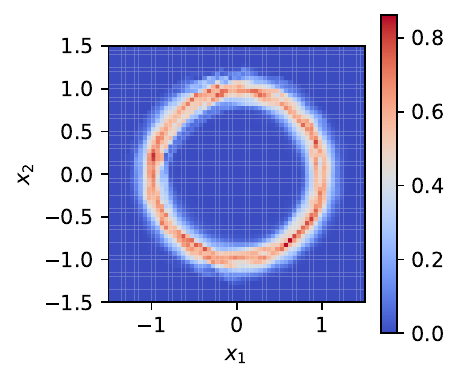}};
        \node (fig3) at (-4, -4) {\includegraphics[width=0.25\textwidth]{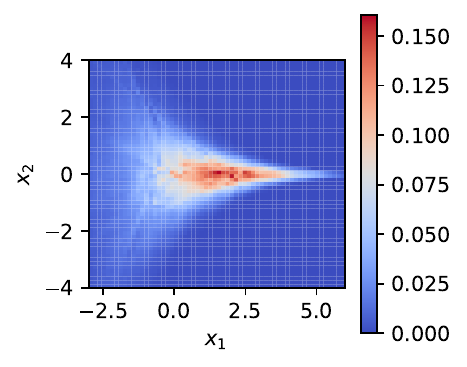}};
        \node (fig4) at (0, -4) {\includegraphics[width=0.25\textwidth]{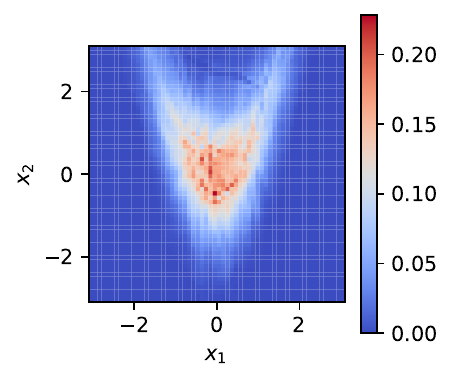}};

    \node (note2) at (-14, -7) {\rotatebox{90}{Exact}};
        \node (fig5) at (-12, -7) {\includegraphics[width=0.25\textwidth]{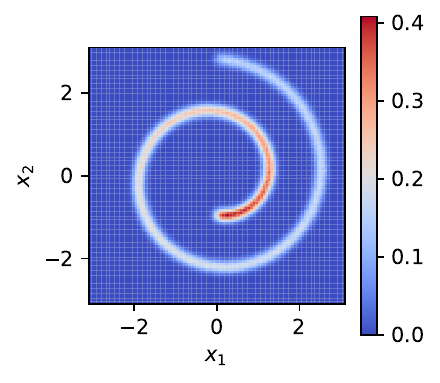}};
        \node (fig6) at (-8, -7) {\includegraphics[width=0.25\textwidth]{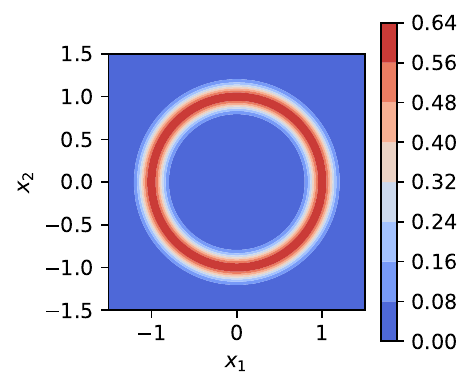}};
        \node (fig7) at (-4, -7) {\includegraphics[width=0.25\textwidth]{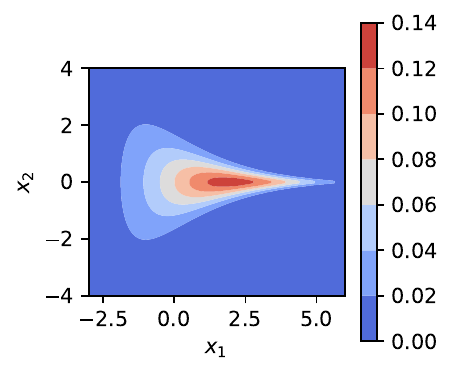}};
        \node (fig8) at (0, -7) {\includegraphics[width=0.25\textwidth]{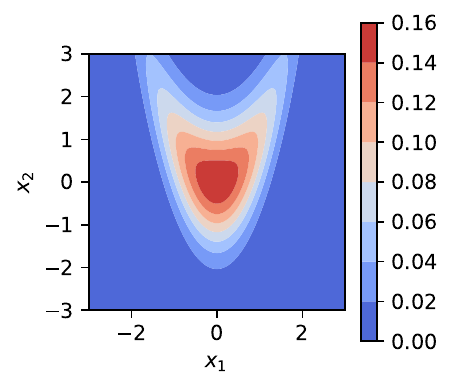}};

        \draw[->, line width=0.4mm] ([yshift=0.1cm]top.south) -- ([yshift=-0.2cm]fig1.north);
        \draw[->, line width=0.4mm] ([yshift=0.1cm]top.south) -- ([yshift=-0.2cm]fig2.north);
        \draw[->, line width=0.4mm] ([yshift=0.1cm]top.south) -- ([yshift=-0.2cm]fig3.north);
        \draw[->, line width=0.4mm] ([yshift=0.1cm]top.south) -- ([yshift=-0.2cm]fig4.north);
    \end{tikzpicture}
\caption{Transport map $M_{Y\rightarrow Z}$ based on paired samples of a five-marginal optimal transport problem, i.e. from the normally distributed one-marginal $Y=[X^{(1)}]$ (top) to a four-marginal output $Z=[X^{(2)},X^{(3)},X^{(4)},X^{(5)}]$ (middle and bottom) consisting of the Swiss roll, ring, funnel, and banana distributions.}
\label{fig:est_post_5marginal}
\end{figure*}

\begin{figure}
    \centering 
    \begin{tabular}{ccc}
    
{\hspace{-0.9cm}    \includegraphics[width=0.38\linewidth]{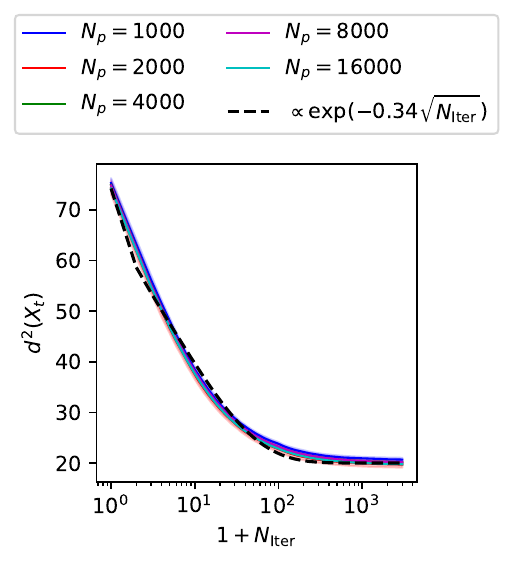}}
    &
    \hspace{-0.6cm}
\includegraphics[width=0.4\linewidth]{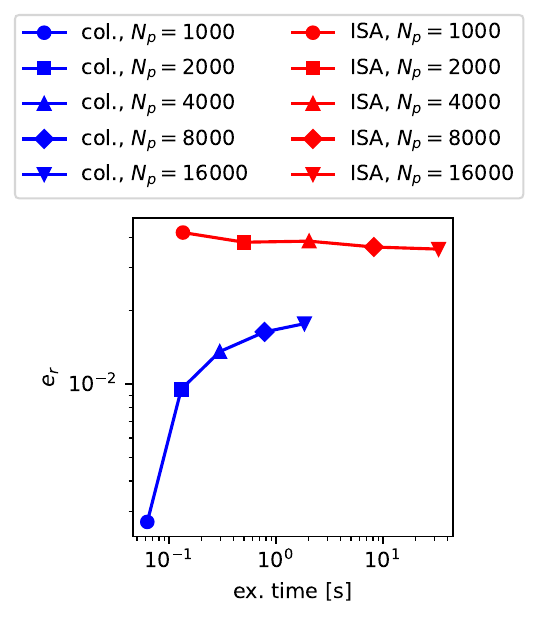}
&
    \hspace{-0.3cm}
    \includegraphics[width=0.34\linewidth]{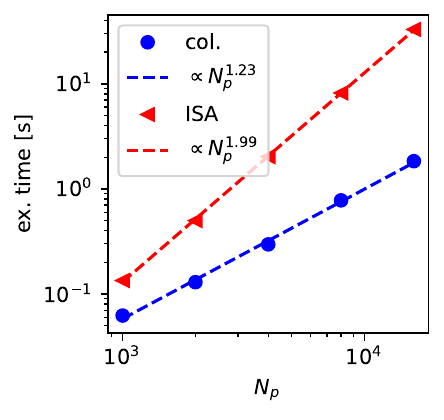}
    \end{tabular}
    \caption{Evolution of the cost function during collision-based MMOT Algorithm~\ref{alg:collOT}  per number of iteration (left), relative error against execution time (middle) and scaling of execution time (right) for a range of $N_p$ in finding the optimal map between five-marginal consisting of normal, Swiss-roll, banana, ring, and funnel densities. In order to have similar orders of magnitude in the relative error, we  compare solution obtained from collisional OT with $1000$ iterations against ISA with $4$ iterations. Here, the solution obtained with ISA using $10$ iterations is considered as the reference solution.}
    \label{fig:5marginal_cost}
\end{figure}

\noindent As an interesting application of MMOT, here we deploy the proposed method to learn a map between normal and four other distributions, i.e. Swiss roll, banana, funnel, and ring, see \cite{baptista2023representation} for details.
\\ \ \\
\noindent First, we take $N_p=2\times 10^4$ samples from five marginals and construct $X=[X^{(1)},...,X^{(5)}]$ where $X^{(1)}\sim \mathcal{N}(0,I)$ and the other marginals follow density of target densities, i.e. Swiss roll, banana, funnel, and ring. We find the optimal map between these five marginals using the proposed Algorithm~\ref{alg:collOT}. The optimal map provides us with the sampling order which is paired to minimize the  transport cost. Then, we train a Neural Network (NN) as a map denoted by $M_{Y\rightarrow Z}$ where the samples of normal distribution $Y:=[X^{(1)}]$ is the input and the other marginals $Z:=[X^{(2)}, X^{(3)},X^{(4)}, X^{(5)}]$ are the output. We construct the NN using $4$ layers, each with $100$ neurons, equipped with $\textrm{tanh(.)}$ as the activation function and a linear operation at the final layer to set the output dimension to $\dim(Z)$. Here, we use Adam's algorithm \cite{kingma2014adam} with a learning rate of $10^{-3}$, take $L^2$ point-wise error between NN estimate and optimally ordered data as the loss function, and carry out $5,000$ iterations to find the NN weights.
\\ \ \\
\noindent For testing, we generate $10^6$ normally distributed samples, i.e. $Y^\mathrm{test} \sim \mathcal{N}( 0, I_{n\times n})$, and feed them as the input into the NN to find $Z^\mathrm{test} = M_{Y \rightarrow Z}(Y^\text{test})$. As shown in Fig.~\ref{fig:est_post_5marginal}, the estimated map via Neural Network trained using optimally paired samples recovers the target densities with a high accuracy. 
\noindent In Fig.~\ref{fig:5marginal_cost}, we analyze the convergence rate of the proposed algorithm, and its cost with respect to execution time and memory consumption. As expected, we see that Algorithm~\ref{alg:collOT} scales linearly with number of samples, the cost of the optimal map decreases exponentially to its optimal value, and  the number of iterations till convergence is not affected by $N_p$. In Appendix~\ref{sec:ot_2by2_5marginal}, we also compare the performance of the proposed collision-based OT Algorithm~\ref{alg:collOT} against ISA, EMD, and Sinkhorn in 2-marginal settings.

\subsection{Optimal map between normal and Swiss roll/banana/funnel/ring density}
\label{sec:ot_2by2_5marginal}

\begin{figure}
    \centering 
    \begin{tabular}{ccccc}
   \hspace{-1cm} \rotatebox{90}{Swiss-Roll and Normal}
    &  
\includegraphics[width=0.25\linewidth]{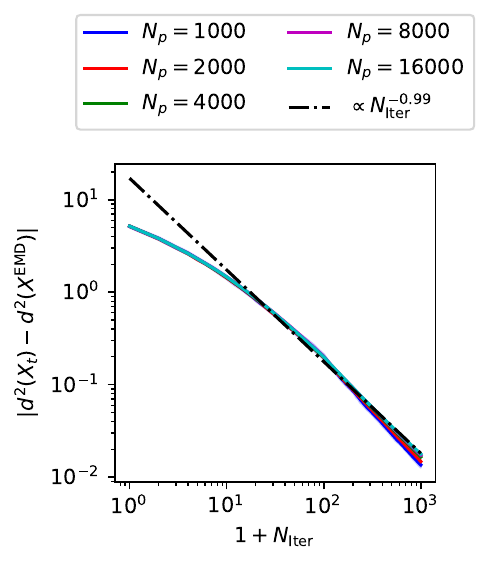}
    & 
\includegraphics[width=0.24\linewidth]{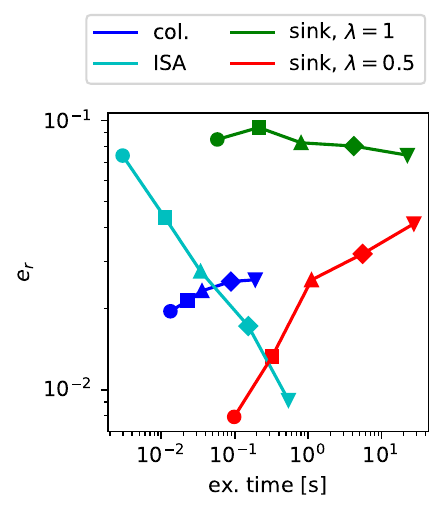}
    &  
\includegraphics[width=0.25\linewidth]{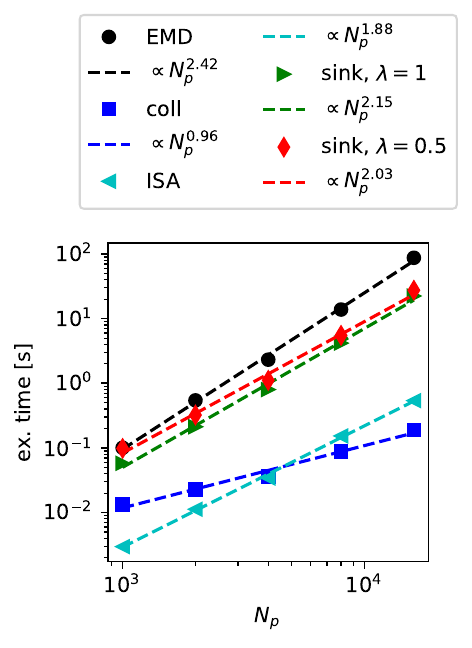}
    & 
\includegraphics[width=0.25\linewidth]{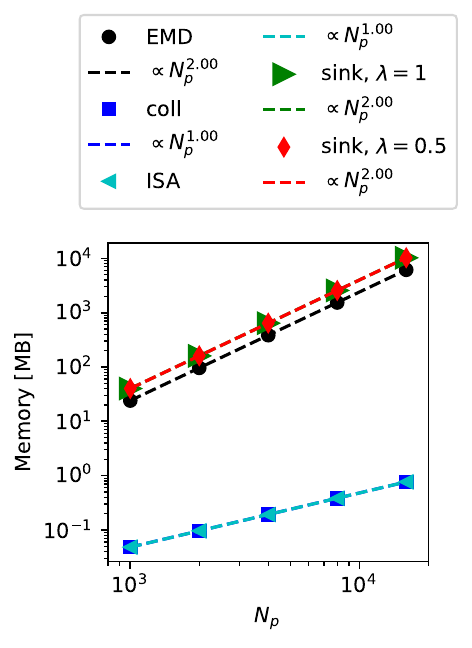}
\\
\hspace{-1cm} \rotatebox{90}{\ \ \ Banana and Normal}
    &
\includegraphics[width=0.25\linewidth]{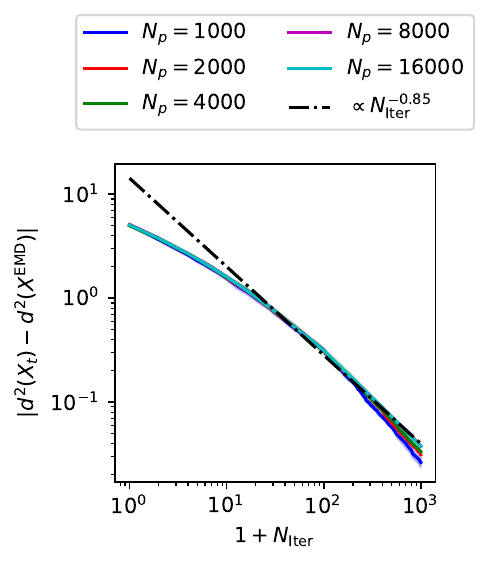}
    &  
\includegraphics[width=0.24\linewidth]{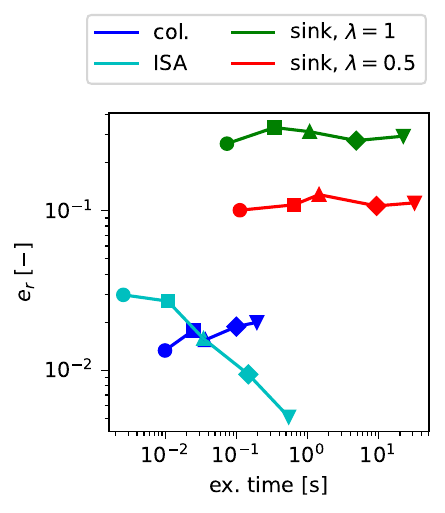}
    &  
\includegraphics[width=0.25\linewidth]{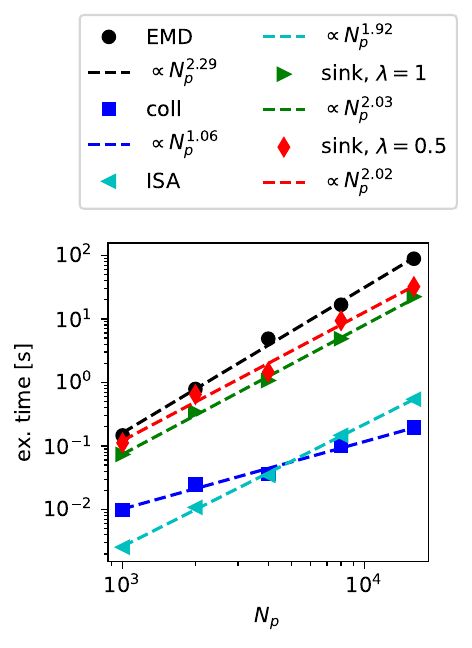}
    &  
\includegraphics[width=0.25\linewidth]{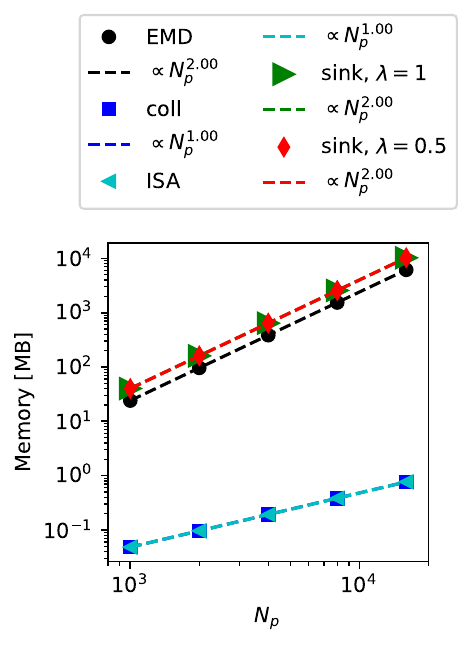}
\\
\hspace{-1cm} \rotatebox{90}{\ \ \ Funnel and Normal}
    &  
\includegraphics[width=0.25\linewidth]{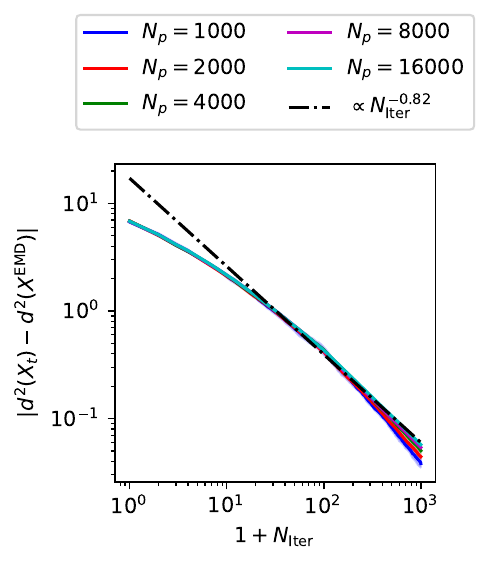}
    &  
\includegraphics[width=0.25\linewidth]{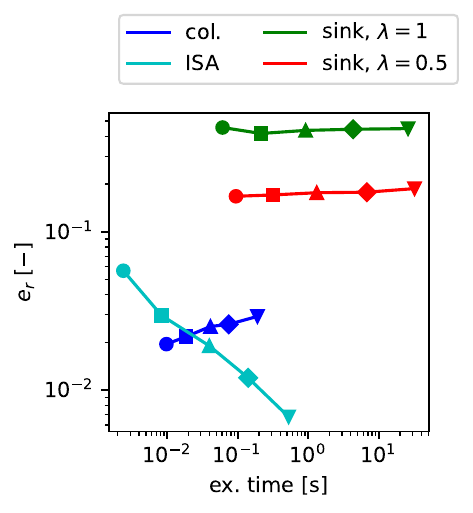}
    &  
\includegraphics[width=0.25\linewidth]{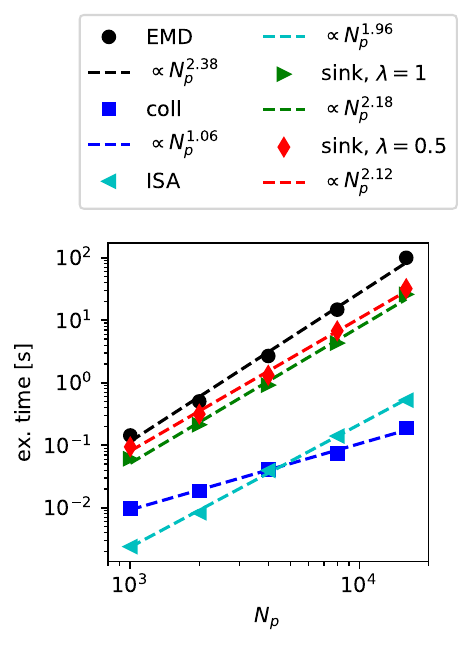}
    &  
\includegraphics[width=0.25\linewidth]{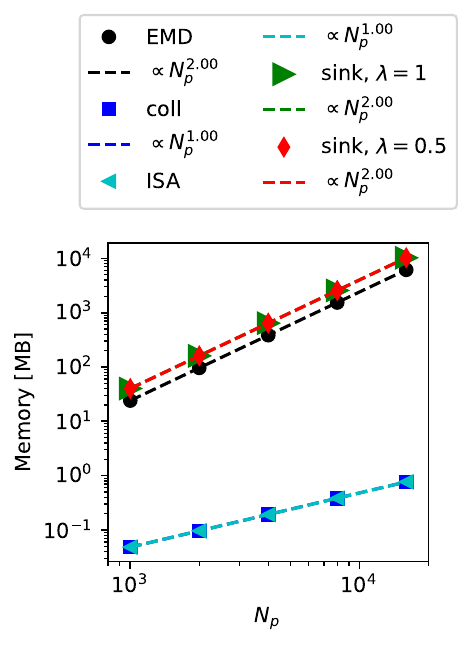}
\\ 
\hspace{-1cm} \rotatebox{90}{\ \ \ \ Ring and Normal}
    &  
\includegraphics[width=0.25\linewidth]{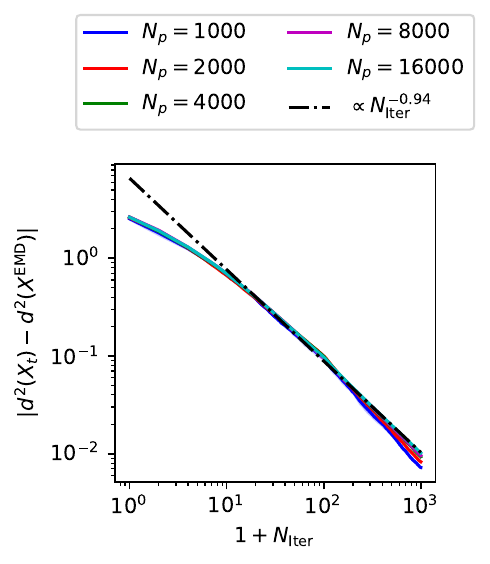}
&  
\includegraphics[width=0.25\linewidth]{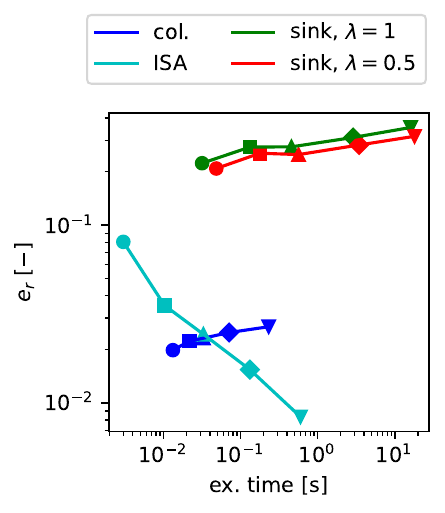}
    &  
\includegraphics[width=0.25\linewidth]{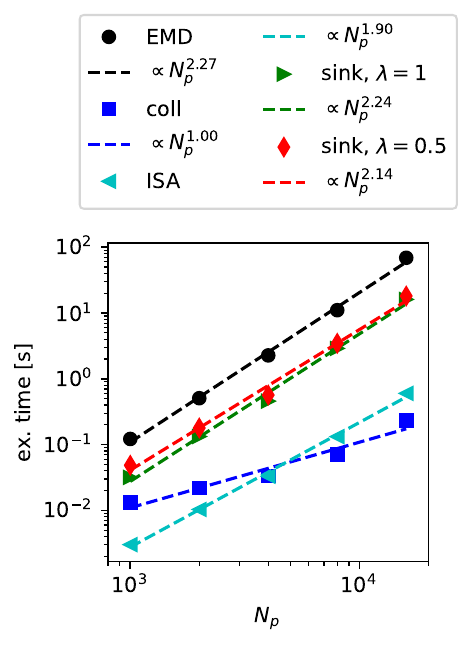}
    &  
\includegraphics[width=0.25\linewidth]{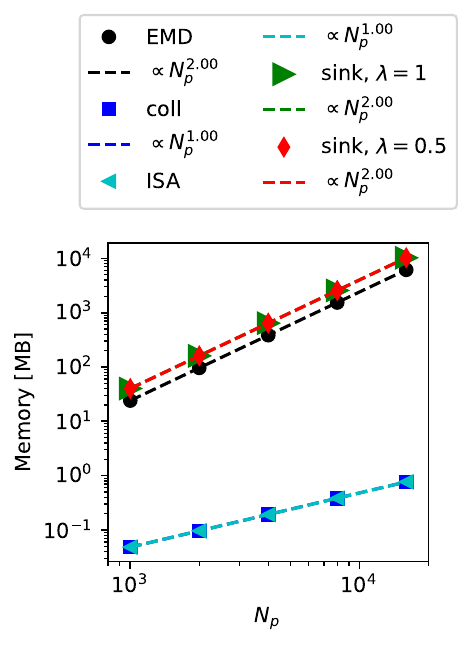}
\\
& (a) & (b) & (c) & (d) 
    \end{tabular}
    \caption{
    Evolution of the error in the Wasserstein distance $d^2(.,.)$ between the proposed collision-based solution algorithm~\ref{alg:collOT} and  EMD (a) relative error in Wasserstein distance using EMD as the reference solution versus execution time (b), scaling of execution time (c), and memory consumption (d) for a range of $N_p$ in finding optimal two-marginal map between normal-Swiss roll, normal-Banana, normal-Funnel, and normal-Ring distribution compared to ISA (one full step), EMD, and Sinkhorn with regularization factor $\lambda=1$ and $0.5$. In (b), the symbols $\circ$, $\square$, $\bigtriangleup$, $\Diamond$, $\triangledown$ correspond to $N_p=1000,2000,4000,8000,16000$, respectively.}
\label{fig:2d_normal_all_other_marginals}
\end{figure}

\noindent Consider two-marginal optimal map between normal distribution $\mu_1=\mathcal{N}(0,I_{2\times 2})$ and a target distribution $\mu_2$. Here, we consider Swiss roll, banana, funnel, and ring as target densities, see \cite{baptista2023representation} for details. We draw $N_p$ samples from the two marginals, and solve the optimal transport problem using the proposed collision-based algorithm \ref{alg:collOT}, EMD and Sinkhron. As shown in Fig.~\ref{fig:2d_normal_all_other_marginals}, the proposed algorithm outperforms the benchmark at a reasonable error, both in terms of execution time and memory consumption.

Furthermore, we have carried out a convergence study by comparing ISA to the proposed collisional method in terms of relative error against complexity and execution time.

\begin{table}[H]
\begin{tabular}{|c|c|c|c|c|c|}
\hline
\multicolumn{3}{|c|}{\textbf{ISA}} & \multicolumn{3}{|c|}{\textbf{Collisional OT}}\\
\hline
No. Coll. & Time [s] & Rel. Error [-] & No. Coll. & Time [s] & Rel. Error [-] \\
\hline
$N_p^2/2$ &
\textbf{0.130 ± 0.001} &
\textbf{1.731e-02 ±1.415e-03} &
500$(N_p/2)$ &
{0.038 ± 0.003} &
{4.933e-02 ± 1.003e-03}
\\ \hline 
2$(N_p^2/2)$ &
0.264 ± 0.007 &
1.609e-03 ± 1.828e-04 &
1000$(N_p/2)$ &
{0.073 ± 0.001}&
{2.533e-02 ± 8.453e-04}
\\ \hline 
3$(N_p^2/2)$ &
0.396 ± 0.016 &
1.038e-03 ± 1.300e-04 &
1500$(N_p/2)$ &
\textbf{0.098 ± 0.004} &
\textbf{1.665e-02 ± 5.533e-04} 
\\ \hline 
4$(N_p^2/2)$ &
0.515 ± 0.017 &
8.999e-04 ± 1.312e-04 &
2000$(N_p/2)$&
0.141 ± 0.006 &
1.236e-02 ± 4.971e-04 
\\ \hline 
5$(N_p^2/2)$ &
0.641 ± 0.007 &
8.613e-04 ± 1.528e-04 &
10000$(N_p/2)$&
0.717 ± 0.043 &
2.418e-03 ± 1.178e-04 
\\ \hline
\end{tabular}
\caption{Evolution of relative error and the execution time in finding the Wasserstein distance between Swiss-Roll and Normal distribution given $N_p=8000$ samples repeated $20$ times using ISA and collisional OT.}
\end{table}

\begin{table}[H]
\begin{tabular}{|c|c|c|c|c|c|}
\hline
\multicolumn{3}{|c|}{\textbf{ISA}} & \multicolumn{3}{|c|}{\textbf{Collisional OT}}\\
\hline
No. Coll. & Time [s] & Rel. Error [-] & No. Coll. & Time [s] & Rel. Error [-] \\
\hline
$N_p^2/2$ &
\textbf{0.146 ± 0.021}&
\textbf{9.164e-03 ±3.611e-04}&
500$(N_p/2)$ &
{0.037 ± 0.003}&
{3.706e-02 ± 4.738e-04}
\\ \hline 
2$(N_p^2/2)$ &
0.287 ± 0.020 &
9.972e-04 ± 1.017e-04 &
1000$(N_p/2)$ &
{0.066 ± 0.001} &
{1.927e-02 ± 8.256e-05} 
\\ \hline 
3$(N_p^2/2)$ &
0.423 ± 0.020&
5.659e-04 ± 6.927e-05&
1500$(N_p/2)$ &
\textbf{0.114 ± 0.010} &
\textbf{1.239e-02 ± 7.734e-04}
\\ \hline 
4$(N_p^2/2)$ &
0.529 ± 0.006 &
5.366e-04 ± 2.189e-05 &
2000$(N_p/2)$&
{0.149 ± 0.006} &
{9.247e-03 ± 3.580e-04} 
\\ \hline 
5$(N_p^2/2)$ &
0.651 ± 0.017&
5.109e-04 ± 9.676e-05 &
10000$(N_p/2)$&
0.699 ± 0.010&
1.820e-03 ± 1.494e-04
\\ \hline
\end{tabular}
\caption{Evolution of relative error and the execution time in finding the Wasserstein distance between Banana and Normal distribution given $N_p=8000$ samples repeated $20$ times using ISA and collisional OT.}
\end{table}

\begin{table}[H]
\begin{tabular}{|c|c|c|c|c|c|}
\hline
\multicolumn{3}{|c|}{\textbf{ISA}} & \multicolumn{3}{|c|}{\textbf{Collisional OT}}\\
\hline
No. Coll. & Time [s] & Rel. Error [-] & No. Coll. & Time [s] & Rel. Error [-] \\
\hline
$N_p^2/2$ &
\textbf{0.155 ± 0.025}&
\textbf{1.497e-02 ± 7.107e-04}&
500$(N_p/2)$ &
{0.042 ± 0.013}&
{5.339e-02 ± 1.043e-03}
\\ \hline 
2$(N_p^2/2)$ &
0.283 ± 0.013&
1.495e-03 ± 5.554e-05&
1000$(N_p/2)$ &
 {0.078 ± 0.006}&
{2.791e-02 ± 1.100e-03}
\\ \hline 
3$(N_p^2/2)$ &
0.400 ± 0.014&
9.913e-04 ± 4.946e-05&
1500$(N_p/2)$ &
\textbf{0.105 ± 0.001}&
\textbf{1.550e-02 ± 7.087e-04}
\\ \hline 
4$(N_p^2/2)$ &
0.561 ± 0.024&
9.807e-04 ± 4.403e-05&
2000$(N_p/2)$&
0.146 ± 0.001&
1.274e-02 ± 5.250e-04
\\ \hline 
5$(N_p^2/2)$ &
0.669 ± 0.010&
8.894e-04 ± 2.912e-04&
10000$(N_p/2)$&
0.727 ± 0.024&
2.639e-03 ± 1.264e-04
\\ \hline
\end{tabular}
\caption{Evolution of relative error and the execution time in finding the Wasserstein distance between Funnel and Normal distribution given $N_p=8000$ samples repeated $20$ times using ISA and collisional OT.}
\end{table}

\begin{table}[H]
\begin{tabular}{|c|c|c|c|c|c|}
\hline
\multicolumn{3}{|c|}{\textbf{ISA}} & \multicolumn{3}{|c|}{\textbf{Collisional OT}}\\
\hline
No. Coll. & Time [s] & Rel. Error [-] & No. Coll. & Time [s] & Rel. Error [-] \\
\hline
$N_p^2/2$ &
\textbf{0.167 ± 0.001}&
\textbf{1.715e-02 ± 9.159e-04}&
500$(N_p/2)$ &
{0.035 ± 0.001}&
{4.979e-02 ± 1.182e-03}
\\ \hline 
2$(N_p^2/2)$ &
0.277 ± 0.019&
1.431e-03 ± 6.684e-05&
1000$(N_p/2)$ &
{0.074 ± 0.007}&
{2.618e-02 ± 3.668e-04}
\\ \hline 
3$(N_p^2/2)$ &
0.440 ± 0.037&
9.930e-04 ± 7.219e-05&
1500$(N_p/2)$ &
\textbf{0.119 ± 0.008}&
\textbf{1.735e-02 ± 5.948e-04}
\\ \hline 
4$(N_p^2/2)$ &
0.607 ± 0.065&
8.594e-04 ± 9.219e-05&
2000$(N_p/2)$&
0.199 ± 0.055&
1.317e-02 ± 1.767e-04
\\ \hline 
5$(N_p^2/2)$ &
0.676 ± 0.023&
8.967e-04 ± 1.576e-04&
10000$(N_p/2)$&
0.753 ± 0.023&
2.540e-03 ± 1.154e-04
\\ \hline
\end{tabular}
\caption{Evolution of relative error and the execution time in finding the Wasserstein distance between Ring and Normal distribution given $N_p=8000$ samples repeated $20$ times using ISA and collisional OT.}
\end{table}

\subsection{Wasserstein distance in several datasets}
\label{sec:app_wasserstein_dataset}
\begin{figure}[H]
    \centering 
    \begin{tabular}{ccc}
    \includegraphics[width=0.3\linewidth]{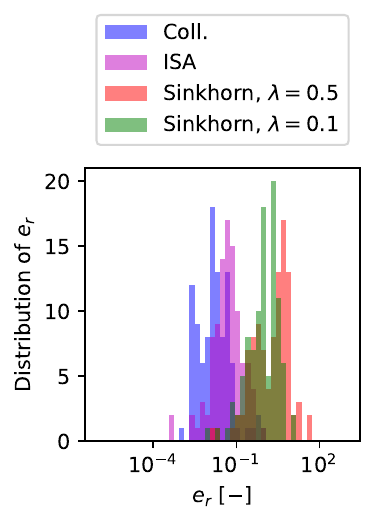}
    &
\includegraphics[width=0.3\linewidth]{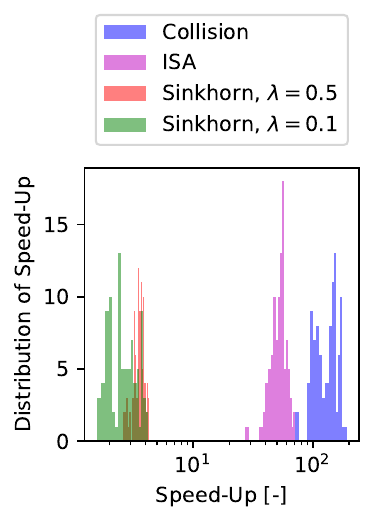}
&
\includegraphics[width=0.3\linewidth]{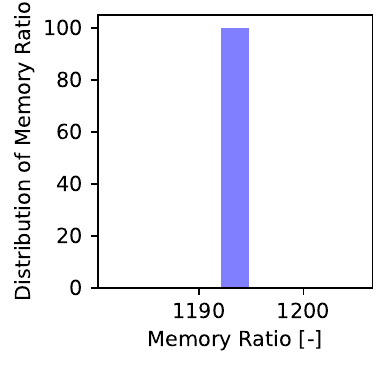}
\\
    \includegraphics[width=0.3\linewidth]{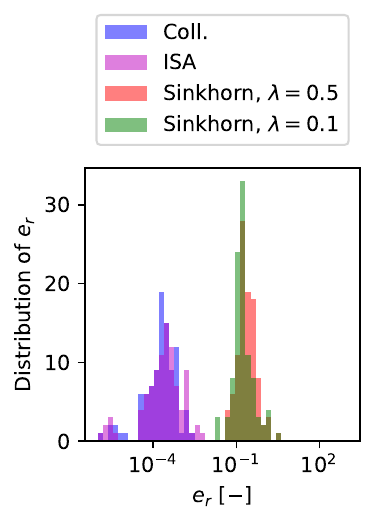}
    &
\includegraphics[width=0.3\linewidth]{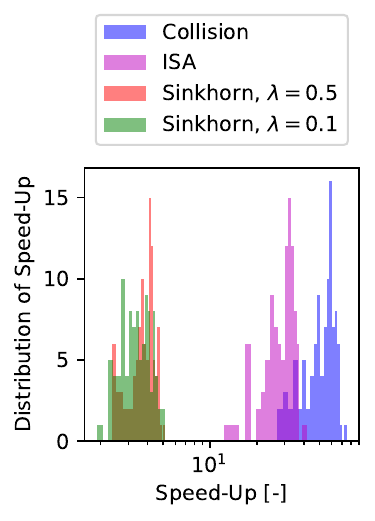}
&
\includegraphics[width=0.3\linewidth]{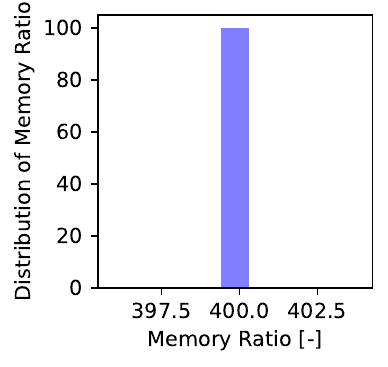}
\\
    \includegraphics[width=0.3\linewidth]{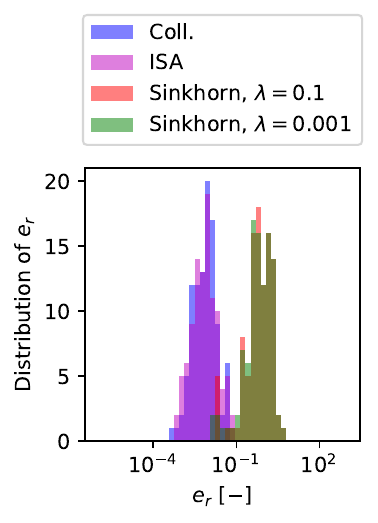}
    &
\includegraphics[width=0.3\linewidth]{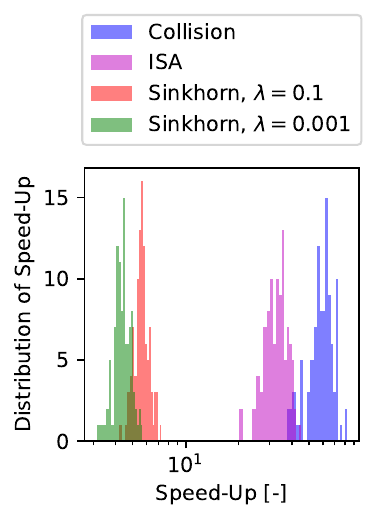}
&
\includegraphics[width=0.3\linewidth]{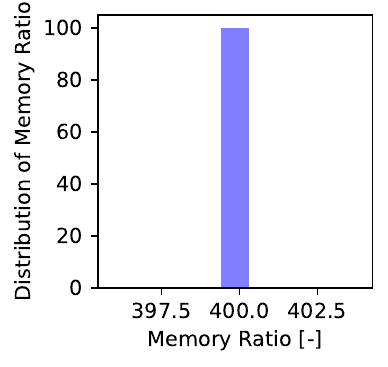}
    \end{tabular}
    \caption{Distribution of relative error $e_r:=|d^2(X^\mathrm{coll})-d^2(X^\mathrm{EMD})|/d^2(X^\mathrm{EMD})$ (left), speed-up, i.e. ratio of EMD execution time to collision method (middle), and ratio of memory consumption, i.e. EMD to collision method, (right) for $100$ randomly selected pairs of picture from JAFFE (top), butterfly (middle), and CelebA (bottom) dataset with randomly selected $2000$ pixel images. Here, we stop ISA after one and collisional OT after $\approx 300$ iterations to be on the same level of relative error, while imposing stopping threshold of $10^{-9}$ for Sinkhorn method to ensure convergence given its large relative error.}
    \label{fig:dist_datasets}
\end{figure}

\noindent As an example from image processing, here we consider optimal transport problem given data from standard datasets such as JAFFE, butterfly, and CelebA. In each case, we take $100$ random pairs of pictures from the dataset, and find the optimal map between them using EMD, Sinkhorn, and the proposed collision-based OT Algorithm~\ref{alg:collOT}. As shown in Fig.~\ref{fig:dist_datasets}, the proposed collision-based approach outperforms Sinkhorn in the distribution of relative error in the Wasserstein distance, speed-up and memory consumption.

\begin{table}[H]
\begin{tabular}{|c|c|c|c|c|c|}
\hline
\multicolumn{3}{|c|}{\textbf{ISA}} & \multicolumn{3}{|c|}{\textbf{Collisional OT}}\\
\hline
No. Coll. & Time [s] & Rel. Error [-] & No. Coll. & Time [s] & Rel. Error [-] \\
\hline
$N_p^2/2$ &
\textbf{0.094 ± 0.014}&
\textbf{2.643e-01 ± 2.006e-01}&
500$(N_p/2)$ &
\textbf{0.022 ± 0.003}&
\textbf{8.456e-03 ± 1.541e-02}
\\ \hline 
2$(N_p^2/2)$ &
0.185 ± 0.010&
1.854e-01 ± 1.348e-01&
1000$(N_p/2)$ &
0.044 ± 0.003&
3.222e-03 ± 4.031e-03
\\ \hline 
3$(N_p^2/2)$ &
0.281 ± 0.032&
6.508e-02 ± 8.337e-02&
1500$(N_p/2)$ &
0.069 ± 0.011&
1.495e-03 ± 1.356e-03
\\ \hline 
4$(N_p^2/2)$ &
0.366 ± 0.017&
2.517e-02 ± 3.309e-02&
2000$(N_p/2)$&
0.089 ± 0.013&
8.556e-04 ± 2.705e-03
\\ \hline 
5$(N_p^2/2)$ &
0.470 ± 0.034&
1.409e-02 ± 1.390e-02&
10000$(N_p/2)$&
0.437 ± 0.029&
1.094e-05 ± 8.197e-06
\\ \hline
\end{tabular}
\caption{Evolution of relative error and the execution time in finding the Wasserstein distance between 100 randomly selected pairs of images from the JAFFE dataset using $N_p=8000$ pixels repeated 20 times using ISA and collisional OT.}
\end{table}

\begin{table}[H]
\begin{tabular}{|c|c|c|c|c|c|}
\hline
\multicolumn{3}{|c|}{\textbf{ISA}} & \multicolumn{3}{|c|}{\textbf{Collisional OT}}\\
\hline
No. Coll. & Time [s] & Rel. Error [-] & No. Coll. & Time [s] & Rel. Error [-] \\
\hline
$N_p^2/2$ &
\textbf{0.217±0.021}&
\textbf{9.796e-03 ±2.636e-02}&
300$(N_p/2)$ &
\textbf{0.031 ± 0.003}&
\textbf{1.028e-02 ± 2.860e-02}
\\ \hline 
2$(N_p^2/2)$ &
0.384 ± 0.031&
8.879e-04 ± 1.596e-04&
500$(N_p/2)$ &
0.054 ± 0.011&
6.254e-03 ± 1.840e-03
\\ \hline 
3$(N_p^2/2)$ &
0.561 ± 0.046&
5.775e-04 ± 6.024e-06&
1000$(N_p/2)$ &
0.103 ± 0.005&
4.561e-03 ± 1.359e-04
\\ \hline 
4$(N_p^2/2)$ &
0.803 ± 0.195&
4.578e-04 ± 4.702e-04&
2000$(N_p/2)$&
0.232 ± 0.059&
2.923e-03 ± 5.096e-03
\\ \hline 
5$(N_p^2/2)$ &
0.987 ± 0.080&
3.681e-04 ± 6.452e-04&
10000$(N_p/2)$&
1.158 ± 0.130&
9.886e-04 ± 1.745e-03
\\ \hline
\end{tabular}
\caption{Evolution of relative error and the execution time in finding the Wasserstein distance between 100 randomly selected pairs of images from the Butterfly dataset using $N_p=8000$ pixels repeated 20 times using ISA and collisional OT.}
\end{table}

\begin{table}[H]
\begin{tabular}{|c|c|c|c|c|c|}
\hline
\multicolumn{3}{|c|}{\textbf{ISA}} & \multicolumn{3}{|c|}{\textbf{Collisional OT}}\\
\hline
No. Coll. & Time [s] & Rel. Error [-] & No. Coll. & Time [s] & Rel. Error [-] \\
\hline
$N_p^2/2$ &
\textbf{0.212 ± 0.014}&
\textbf{4.405e-03 ± 3.455e-03}&
500$(N_p/2)$ &
0.057 ± 0.003&
6.943e-03 ± 3.699e-03
\\ \hline 
2$(N_p^2/2)$ &
0.420 ± 0.027&
3.668e-04 ± 2.785e-04&
1000$(N_p/2)$ &
\textbf{0.120 ± 0.014}&
\textbf{4.643e-03 ± 1.926e-03}
\\ \hline 
3$(N_p^2/2)$ &
0.567 ± 0.038&
5.920e-04 ± 5.400e-04&
1500$(N_p/2)$ &
0.165 ± 0.014&
3.392e-03 ± 0.692e-03
\\ \hline 
4$(N_p^2/2)$ &
0.733 ± 0.039&
5.272e-04 ± 8.864e-04&
2000$(N_p/2)$&
0.212 ± 0.016&
2.229e-03 ± 1.022e-03
\\ \hline 
5$(N_p^2/2)$ &
0.903 ± 0.028&
4.089e-04 ± 3.752e-04&
10000$(N_p/2)$&
1.051 ± 0.041&
1.096e-03 ± 1.018e-03
\\ \hline
\end{tabular}
\caption{Evolution of relative error and the execution time in finding the Wasserstein distance between 100 randomly selected pairs of images from the CelebA dataset using $N_p=8000$ pixels repeated 20 times using ISA and collisional OT.}
\end{table}

\subsection{Coloring images}
\label{sec:ColorImage}

\begin{figure}[H]
    \begin{tikzpicture}
        \node (fig1) at (-4, 3) {\includegraphics[width=0.25\textwidth]{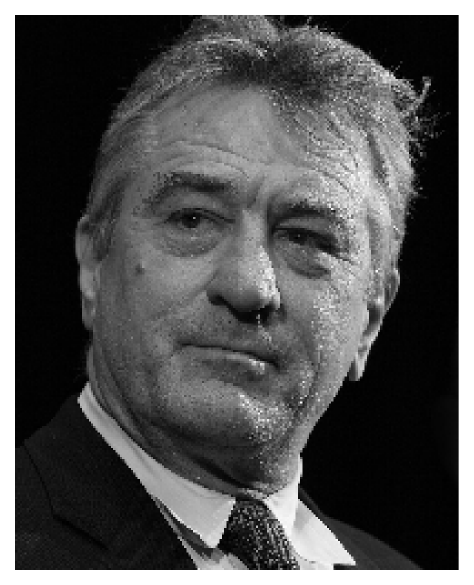}};
        \node (fig2) at (4, 3) {\includegraphics[width=0.25\textwidth]{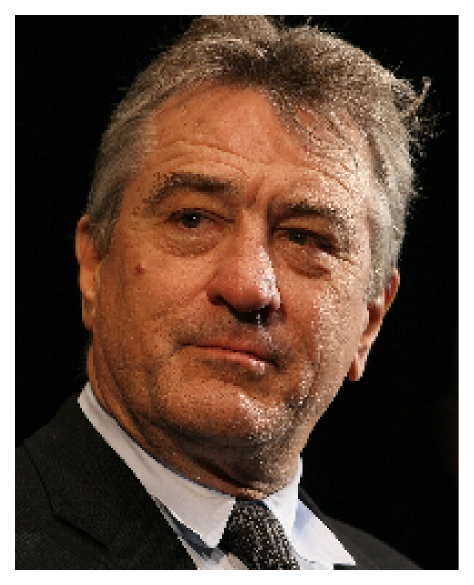}};
        \node (fig3) at (-4, -3) {\includegraphics[width=0.25\textwidth]{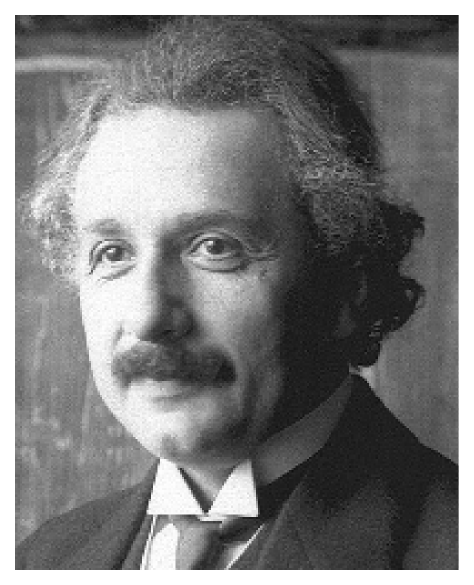}};
        \node (fig4) at (4, -3) {\includegraphics[width=0.25\textwidth]{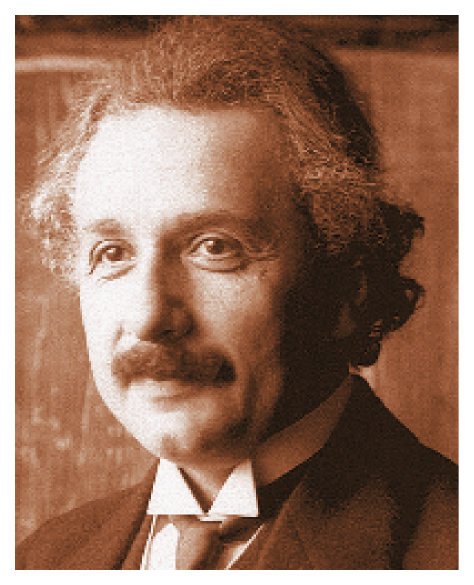}};

        \draw[->, line width=0.5mm] (fig1.east) -- node[midway, above] {$M_{Y\rightarrow Z}(Y^\text{train})$} (fig2.west);
    \draw[->, line width=0.5mm] (fig3.east) -- node[midway, above] {$M_{Y\rightarrow Z}(Y^\text{test})$} (fig4.west);
    \end{tikzpicture}
        \caption{After training a network to learn the map $M_{Y\rightarrow Z}$ on the optimally paired data of gray and colored portrait of Robert De Niro (top), we test the network to find a colored picture of Einstein given a gray portrait of him (bottom).}
    \label{fig:portrait_color_transfer}
    
\end{figure}

\noindent Assume we are given a picture of Robert De Niro\footnote{This image is taken from the public domain available on \url{commons.wikimedia.org/wiki/File:Robert_De_Niro_KVIFF_portrait.jpg}.} in gray and color. We intend to learn the map between the colored and black/white pictures and use it to turn another black/white picture into a colorful one. We take $4,000$ samples from De Niro's picture, use the collision-based algorithm to find the optimal map on discrete points and train a NN denoted by $M$ with $L^2$-pointwise error between optimally sorted data points and NN prediction as loss.  We construct the NN using $4$ layers, each with $100$ neurons, equipped with $\textrm{tanh(.)}$ as activation function and use Adam's algorithm \cite{kingma2014adam} with a learning rate of $10^{-3}$, take $L^2$ point-wise error between NN estimate and optimally ordered data as the loss function, and carry out $5,000$ iterations to find the NN weights. Afterward, we test NN by plugging in a black/white portrait of Albert Einstein\footnote{This image is taken from the public domain available on \url{commons.wikimedia.org/wiki/File:Three_famous_physicists.png}.} as input and recover a colored picture as the output, see Fig.~\ref{fig:portrait_color_transfer}.

\end{document}